\documentclass[10pt,twocolumn,letterpaper]{article}

\usepackage[pagenumbers]{cvpr} 

\usepackage[dvipsnames,table,xcdraw]{xcolor}

\usepackage{graphicx}
\usepackage{amsmath}
\usepackage{amssymb}
\usepackage{booktabs}
\usepackage{multirow}
\usepackage{booktabs}
\usepackage{rotating}
\usepackage{array}
\usepackage{graphicx}
\usepackage{float}
\usepackage{arydshln}
\usepackage{tabularx}
\newtheorem{theorem}{Theorem}
\usepackage{tabularx}
\usepackage{multirow}

\definecolor{cvprblue}{HTML}{FF1493} 
\definecolor{blue}{HTML}{0071BC}
\definecolor{red}{HTML}{ED1C24}

\definecolor{citecolor}{HTML}{0071BC}
\definecolor{linkcolor}{HTML}{ED1C24}
\definecolor{blue}{HTML}{0071BC}
\definecolor{red}{HTML}{ED1C24}
\newtheorem{definition}{Definition}
\newtheorem{lemma}{Lemma}
\usepackage[pagebackref,breaklinks,colorlinks,citecolor=cvprblue]{hyperref}

\def\paperID{11476} 
\def\confName{CVPR}
\def\confYear{2024}

\title{LanDA: Language-Guided Multi-Source Domain Adaptation}

\author{\textbf{Zhenbin Wang}\\
Sichuan University\\
{\tt\footnotesize wangzhenbin@stu.scu.edu.cn}
\and
\textbf{Lei Zhang\thanks{The corresponding author.}, ~Lituan Wang}\\
Sichuan University\\
{\tt\small \{\footnotesize{leizhang,lituanwang}\}@scu.edu.cn}
\and
\textbf{Minjuan Zhu}\\
Sichuan University\\
{\tt\footnotesize zhuminjuan@stu.scu.edu.cn}
}

\begin{document}
\maketitle
\begin{abstract}
Multi-Source Domain Adaptation (MSDA) aims to mitigate changes in data distribution when transferring knowledge from multiple labeled source domains to an unlabeled target domain. However, existing MSDA techniques assume target domain images are available, yet overlook image-rich semantic information. Consequently, an open question is whether MSDA can be guided solely by textual cues in the absence of target domain images. By employing a multimodal model with a joint image and language embedding space, we propose a novel language-guided MSDA approach, termed LanDA, based on optimal transfer theory, which facilitates the transfer of multiple source domains to a new target domain, requiring only a textual description of the target domain without needing even a single target domain image, while retaining task-relevant information. We present extensive experiments across different transfer scenarios using a suite of relevant benchmarks, demonstrating that LanDA outperforms standard fine-tuning and ensemble approaches in both target and source domains.
\end{abstract}    
\vspace{-1.5em}
\section{Introduction}
\vspace{-0.5em}
\label{sec:intro}

Generalizing outside of the training domain is vital for constructing robust computer vision models, and Domain Adaptation (DA) is an efficacious and prevalent technique to achieve this capability. When multiple source domains are available and multiple distribution shifts occur simultaneously, the problem evolves into a more challenging scenario known as Multi-Source Domain Adaptation (MSDA). Conventional MSDA techniques~\cite{montesuma2021wasserstein, ahmed2021unsupervised, wang2022metateacher} rely on visual backbones, often necessitating a substantial volume of additional target domain image data to acquire the target domain's probability distribution. 
In many settings, obtaining target domain images may pose challenges. Collecting data for an uncommon target domain is frequently a time-consuming and costly, yet describing them linguistically can be relatively straightforward (\eg, domains characterized like \textit{nighttime}, \textit{fog}, \textit{painting}, and \textit{infrared light}). Therefore, we propose a novel and \textit{unexplored} problem: how to improve target domain performance using solely language and multiple source domain images, absence any target domain images.

Visual-Language Foundational Models (VLFMs), such as CLIP~\cite{radford2021learning}, ALING~\cite{jia2021scaling}, Flamingo~\cite{alayrac2022flamingo}, BLIP~\cite{li2022blip, li2023blip}, present significant potential for using language to guide visual tasks by aligning text and images within shared embedding spaces. However, the effective utilization of these powerful foundational models is non-trivial in design.
Moreover, recent research~\cite{zhang2021domain} has show that the backbone network of VLFMs significantly outperform many backbone networks exclusively trained on images, including ResNet~\cite{he2016deep}, visual transformer~\cite{dosovitskiy2020image}, and big transfer~\cite{kolesnikov2020big}. Indeed, VLFMs have gradually become a commonly used feature extraction network in computer vision~\cite{singha2023ad}. While VLFMs demonstrate impressive generalization capabilities, zero-shot classifiers derived from them frequently yield significantly poorer performance compared to models specifically trained for downstream tasks~\cite{kumar2022fine, radford2021learning}. When multiple source domain datasets are accessible, a common practice is to fine-tune models on the source domains data. While this enhances accuracy on the source domains, it often diminishes performance on unseen target domains.


Currently, only limited works explored VLFMs for distribution migration problems. While given that full parameter fine-tuning may distort the favorable attributes of pre-trained features~\cite{kumar2022fine}, prior research has primarily followed two main routes, \textit{i.e.}, designing template prompts then adapting prompts to suit downstream tasks~\cite{zhou2022learning, zhou2022conditional, bose2023stylip, singha2023ad}, and implementing a linear probe on the features then fine-tuning the model's backbone~\cite{dunlap2022using}. Manual prompt template design lacks clarity and scalability, rendering it unsuitable for widespread adoption. In contrast, linear probing has been shown to yield more resilient classifiers~\cite{radford2021learning, kumar2022fine}. However, those methods primarily focus on single-source domain settings.

In this work, we present \textbf{LanDA}, a language-guided MSDA method that utilizes CLIP or similar models as the backbone network, harnessing their extensive domain-level knowledge (\eg, \textit{realistic}, \textit{painting} and \textit{sketch}) within a shared image-language embedding space. Our approach requires only source domain images and textual descriptions of both the target and source domains (\eg, the description of source domains is \textit{a realistic photo of a [CLS]}, \textit{a painting of a [CLS]}, and the target domain is \textit{a sketch of a [CLS]}).

Specifically, we freeze CLIP's parameters and incorporate lightweight \textit{domain-specific augmenters} (composed of multi-layer perceptron) into the image embedding for each source domain, the resulting augmentered embeddings are termed as \textit{extended domains}. To align the distribution and obtain domain-invariant features, we introduce a method for computing the Wasserstein cost matrix within a visual-language model that includes both image and text information. While minimizing the Wasserstein distance between each extended domain, the inter-class distance of the text embedding space is considered. Finally, the convex combination of joint distributions from multiple extended domains not only retain the class-specific information of the source domains while eliminating class-irrelevant information.   
We support this novel conceptual approach by deriving a generalization bound on the target error. 

Our main contributions are (1) the \textit{first} introduction of the MSDA with Language problem, (2) proposing LanDa, a novel language-guided MSDA method that leverages VLFMs to achieve domain adaptation without any target domain image data, (3) we propose a cost function suitable for VLFMs, and evaluate our method across several benchmarks, the results show that LanDA achieves the highest accuracy in both target and source domain samples.

\vspace{-0.4em}
\section{Related Work}
\vspace{-0.6em}
\label{sec:bg}

\paragraph{Multi-Source Domain Adaptation (MSDA).}
MSDA address unsupervised domain adaptation by employing source dataset containing multiple domains, so it should consider both domain divergences among various sources and domain shifts between the source and target. MSDA mainly have two strategies, i.e., statistical discrepancy combination~\cite{mansour2009domain, hoffman2018algorithms}, and cross-domain feature extraction~\cite{venkat2020your, mancini2018boosting}. 
The first strategy computes the statistical discrepancy between each source domain and target domain, and then combines all predictions. For pairwise training, methods such as adversarial learning~\cite{xu2018deep, zhao2020multi} and moment matching~\cite{peng2019moment} have been utilized. For determining weight assignments, various techniques are commonly used, including weighted averages~\cite{peng2019moment}, perplexity scores~\cite{xu2018deep} and Wasserstein distances~\cite{zhao2020multi}. The second strategy employing multiple classifiers while sharing one feature extractor, all domain distributions are aligned implicitly. 
However, these conventional methods make the assumption that the target domain image is readily available and depend on the visual backbone. 
Unlike these, our work assumes no access to any target domain image data, with only multiple labeled source domains available. Our approach accomplishes domain adaptation through language-based guidance and adopts large VLFMs like CLIP as the backbone.

\paragraph{Visual-Language Foundational Models (VLFMs).} In recent years,
with the advancement of transformer for both vision~\cite{dosovitskiy2020image} and language~\cite{vaswani2017attention} tasks, large-scale pre-trained models~\cite{wei2022emergent, radford2021learning, brown2020language, chowdhery2022palm, jia2021scaling, touvron2023llama} have gained popularity. They serve as valuable, task-agnostic foundational tools for various downstream applications in computer vision and natural language processing, consistently set new state-of-the-art benchmarks across multiple tasks. VLFMs bridge the gap between images and text by utilizing a shared embedding space to achieve cross-modal learning. Various pre-training schemes have been adopted in VLFMs, including contrastive learning~\cite{radford2021learning, lu2019vilbert}, masked language modeling~\cite{su2019vl}, and masked region modeling~\cite{chen2019uniter}. While VLFMs demonstrate great potential, their effectiveness for domain adaptation remains to be comprehensively investigated.\\
\indent Numerous vision models are pretrained on large-scale datasets like ImageNet to obtain initialized parameters, and all parameters are updated during transfer learning. In the case of large-scale VLFMs, we can still obtain the network's initialization parameters from the image encoder. However, updating all parameters can lead to a disparity between image embedding and text embedding. This misalignment adversely affects the applicability of downstream tasks.
As one of the most representative VLFMs, CLIP~\cite{radford2021learning} has been rarely leveraged for domain generalization or domain adaptation, with existing efforts~\cite{dunlap2022using, cho2023promptstyler, yu2023open, shu2023clipood} utilized fine-tuning techniques and confined to single-source settings. Our work aims to advance the capability of CLIP for language-guided MSDA. To our best knowledge, this direction has remained unexplored in prior arts.
\vspace{-1.3em}
\paragraph{Optimal Transport (OT).}
OT theory~\cite{kantorovich1942transfer, peyre2019computational} defines distance metrics between probability distributions, furnishing a theoretical framework to investigate optimal transport plans for mapping one distribution onto another.
Having matured as a mathematical field, OT has recently been proposed as a probabilistic framework in machine learning for comparing and manipulating probability distributions~\cite{montesuma2023recent}. OT is also known by names such as the Wasserstein distance, Dudley metric, Kantorovich metric, or earth mover's distance.\\
\indent 
The cost function is a fundamental mental concept in OT, as it quantifies the distance between samples from distinct distributions. Compared to other distance metrics, the Wasserstein distance captures the geometric structure between distribution supports and exhibits greater sensitivity to feature distribution shifts across domains. This provides a mathematically grounded distance measure, with notable empirical performance. 
In machine learning, the Wasserstein distance serves both as an adaptation loss function, minimizing the distance to align distributions~\cite{long2018conditional, chen2019transferability}, and as a means to map features from different domains into a Wasserstein space for adaptive classification~\cite{zhou2020bbn, courty2017joint}. 
However, existing cost functions are inadequate for VLFMs due to their failure to account for the text embeddings.
\begin{figure*}[htbp]
	\centering
	\includegraphics[width=0.99\linewidth]{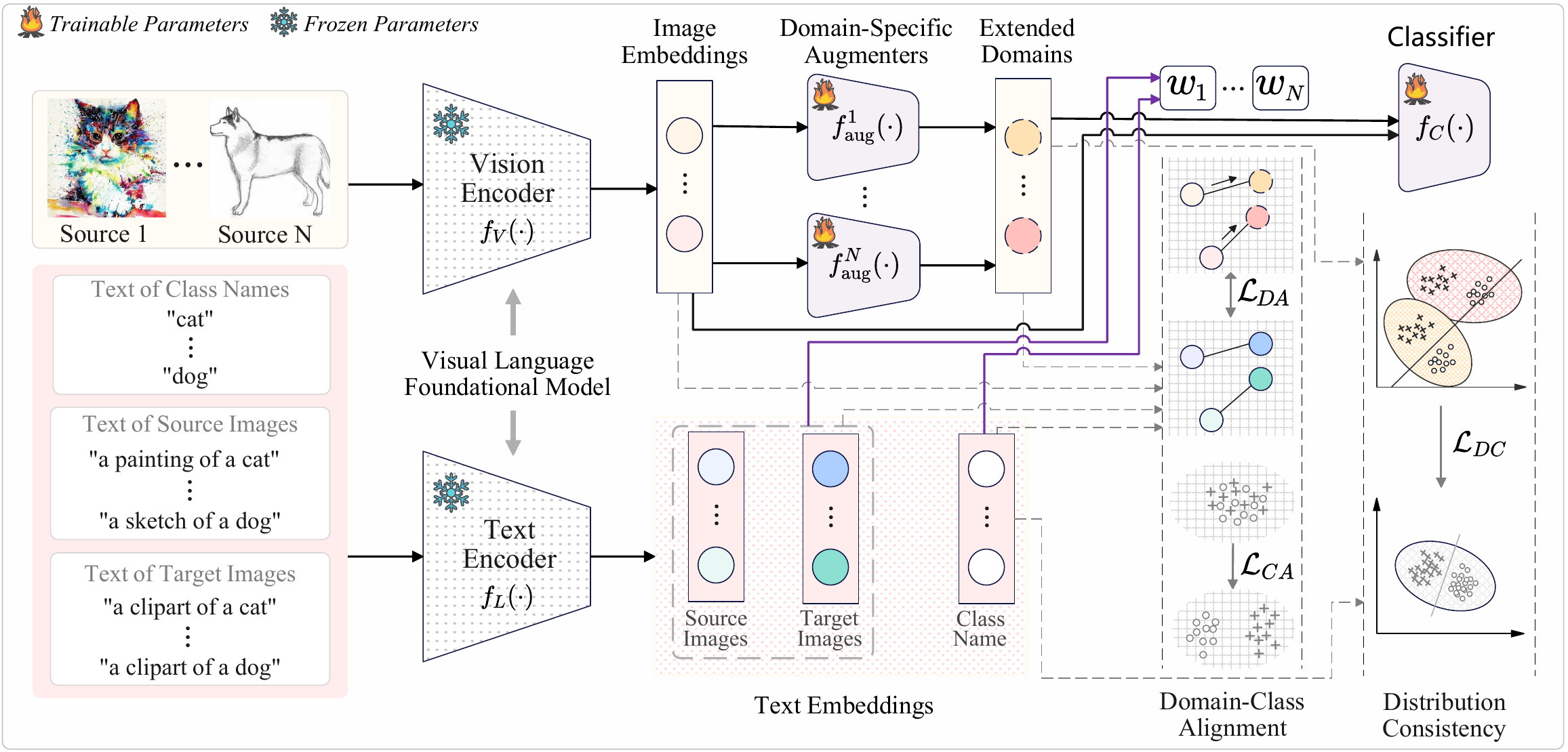}
	\vspace{-0.4em}
	\caption{\textbf{An overview of the proposed two-stage framework.} The training dataset comprises images from multiple source domains and various categories. In the absence of any image samples from the target domain, our goal is to generalize our model to the unseen \textit{clipart} domain. To achieve this, we employ VLFMs and learn domain-level knowledge from the text. Each domain-specific augmenter utilizes the domain-class alignment loss to align the image embeddings from the source domain to the unseen target domain, while preserving their class information. To effectively leverage the knowledge from multiple source domains, we propose a cost matrix function that projects the extended domains and text embeddings of class name into the Wasserstein space to learn domain-invariant information.}
	\label{fig:frameworks}
	\vspace{-0.7em}
\end{figure*}

\section{Language-Guided MSDA}
\vspace{-0.5em}
We tackle the MSDA problem utilizing solely the linguistic descriptions of the multiple source domains and the foreseen but unseen target domain. More formally,

there are $N$ different underlying source distributions denoted as $\{Q_{k}\}_{k=1}^{N}$, and one underlying target distribution denoted as $\{P\}$. It is important to note that $Q_{k}$ is not equal to $P$, leading to domain shift. Additionally, we have the following data:
\vspace{0.1em}
\begin{itemize}
	\item Visual-Language Foundational Models $\{f_{V}, f_{L}\}$, with text encoder $f_{L}$ and visual encoder $f_{V}$.
	\item Image-label pair
	$\mathcal{D}^{(Q_{k})} = \{(x_{i}^{(Q_{k})},y_{i}^{(Q_{k})}  )  \}_{i=1}^{n_{Q_{k}}} $
	from multiple source domains, where $n_{Q_{k}}$ denotes the number of samples in the $k$-th source domain.
	\item The domain descriptions $\{ \mathcal{T}_{Q_{k}} \}_{k=1}^{N}$ of the multiple source domains, and domain description $\{ \mathcal{T}_{P} \}$ of the target domain. Let $\{\{\mathcal{T}_{C}^{k,i}\}_{i=1}^{n_{Q_k}}\}_{k=1}^{N}$ to represent the class name. Furthermore, $\mathcal{T}_{Q_{k}} \! \circ \! \mathcal{T}_{C}^{k,i}$ and $\mathcal{T}_{P} \! \circ \! \mathcal{T}_{C}^{k,i}$ denote the composition of the domain name and the class name. For example, if $\mathcal{T}_{Q_k} = $ \textit{painting}, $\mathcal{T}_{C}^{k,i} = $ \textit{cat}, the composition $\mathcal{T}_{Q_k} \! \circ \! \mathcal{T}_{C}^{k,i} = $ \textit{a painting of a cat}. 
\end{itemize}
\vspace{0.1em}
Our approach can be decomposed into two stages. The first stage entails learning multiple \textit{domain-specific augmenters} to "place" each source domain image embedding into its respective \textit{extended domain}. This serves to (1) align with the unseen target domain, and (2) remove source domain class-irrelevant information, retaining only domain-invariant features. The second stage performs linear probing on the expanded domains and image embeddings to generate a linear classifier. Fig.\ref{fig:frameworks} presents an overview of the proposed framework.
\vspace{-1.0em}
\subsection{Training Domain-Specific Augmenters}
\vspace{-0.5em}

We choose CLIP~\cite{radford2021learning} as our visual-language foundational model. For a given $k$-th source domain image $x_{i}^{(Q_{k})}$ and its corresponding text prompt $\mathcal{T}_{Q_k} \! \circ \! \mathcal{T}_{C}^{k,i}$, $f_V(x_{i}^{(Q_{k})})$ and $f_L(\mathcal{T}_{Q_k} \! \circ \! \mathcal{T}_{C}^{k,i})$ denote the image embedding and text embedding respectively.

\begin{figure*}[htbp]
	\centering
	\includegraphics[width=\linewidth]{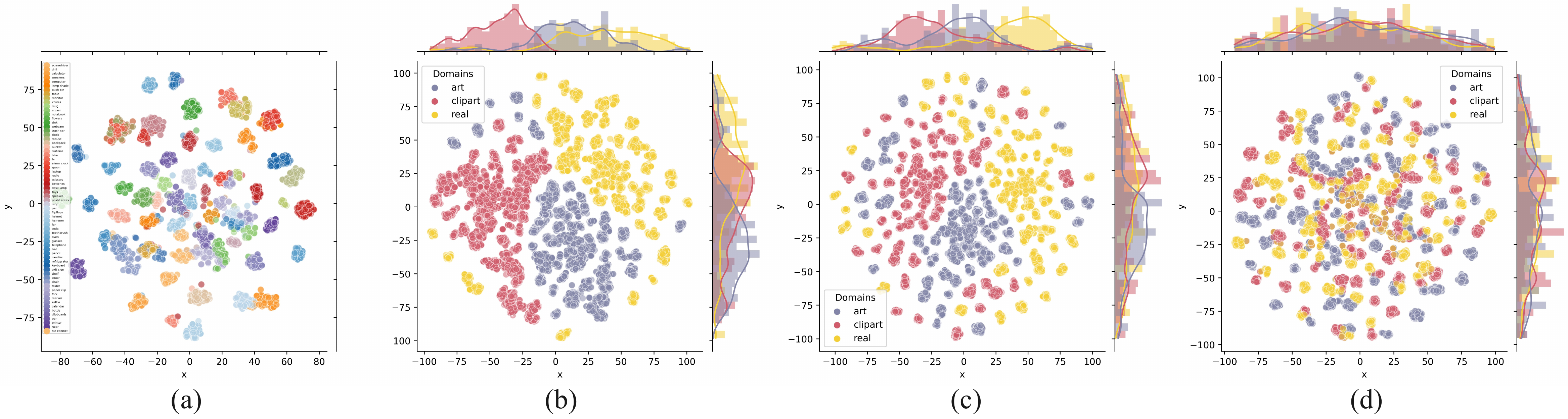}
	\vspace{-1em}
	\caption{\textbf{t-SNE visualizations} on the Office-Home dataset for transfer task A,C,R→P. In figure (a), the image embeddings of CLIP are depicted, with different colors indicating distinct classes. The remaining figure represent the extended domains, where each color corresponds to a different extended domain image embeddings. Three approaches to train domain-specific augmenters are compared: (b) using only $\mathcal{L}_{DA}$; (c) using $\mathcal{L}_{DA}$ and $\mathcal{L}_{CA}$; and (d) using  $\mathcal{L}_{DA}$, $\mathcal{L}_{CA}$ and $\mathcal{L}_{DC}$.}
	\label{fig:onecol}
	\vspace{-1em}
\end{figure*}

The initial stage of LanDA is dedicated to the acquisition of domain augmented network, which is composed by $N$ \textit{domain-specific augmenters} $\{f_{\textrm{aug}}^{k}\}_{k=1}^N$. Each $f_{\textrm{aug}}^{k}$ transform the image embedding $f_V(x_{i}^{(Q_{k})})$ into its respective \textit{extended domain}.
It is crucial to highlight that these domain-specific augmenters do not share weights. Nonetheless, the decision boundaries among the various extended domains may not align perfectly (refer to Fig.\ref{fig:onecol}(c)). We posit that this discrepancy arises from variations in the probability distributions among the different extended domains, thus necessitating the imposition of constraints on the domain-specific augmenters. Consequently, we train $f_{\textrm{aug}}^{k}$ using a combination of two losses: \textit{Domain-Class Alignment} and \textit{Distribution Consistency}.

\paragraph{Domain-Class Alignment.} 
The domain alignment loss utilizes the difference between the source and target domain text embeddings to steer the source domain image embeddings toward the target domain image embeddings.
Some VLFMs such as CLIP employs cosine similarity to quantify the correlation between images and text. The model is trained by maximizing the cosine similarity within the embedding space of similar images and text, while minimizing it within the embedding space of unrelated images and text. Although CLIP establishes a correspondence between the image embeddings and the text embeddings, the precise mapping mechanism remains elusive, making it challenging to describe the mapping relationship between them with a specific function. 
Previous works~\cite{patashnik2021styleclip, gal2021stylegan} assumed the existence of a "global space" in which the direction from the source to target domain text embeddings aligns with the source to target domain image embeddings. We follow this assumption. Consequently, we train the $k$-th domain-specific augmenter using
\setlength{\abovedisplayskip}{5pt}
\setlength{\belowdisplayskip}{5pt}
\begin{eqnarray}\label{eq:eq1}\small
	\begin{split}
		\mathcal{L}_{DA}(f_{\mathrm{aug}}^k)=\sum_{i=1}^{n_{Q_k}} 1-(\frac{f_{\mathrm{aug}}^k(f_V(x_{i}^{(Q_{k})}))-f_V(x_{i}^{(Q_{k})})}{ \| f_{\mathrm{aug}}^k(f_V(x_{i}^{(Q_{k})}))-f_V(x_{i}^{(Q_{k})})  \| }\\
		\cdot \frac{f_L(\mathcal{T}_{P} \! \circ \! \mathcal{T}_{C}^{k,i})-f_L(\mathcal{T}_{Q_k} \! \circ \! \mathcal{T}_{C}^{k,i})}{ \| f_L(\mathcal{T}_{P} \! \circ \! \mathcal{T}_{C}^{k,i})-f_L(\mathcal{T}_{Q_k} \! \circ \! \mathcal{T}_{C}^{k,i}) \| } ). 
	\end{split}
\end{eqnarray}

Although Eq.(\ref{eq:eq1}) aligns the source domain image embeddings towards the unseen target domain, prior work~\cite{dunlap2022using} indicates that optimizing $\mathcal{L}_{DA}$ degrades class-specific representation, resulting in extended domains that fail to retain class-specific traits and exhibit only minor differences of distinct images.
Fig.\ref{fig:onecol}(b) displays the visualization of the training results solely obtained using Eq.(\ref{eq:eq1}). It is evident that numerous clusters are closely situated and lack distinct separability, suggesting a limited classification capability of the model. Fortunately, 
by leveraging CLIP's zero-shot capacity to accurately classify image categories, we can add an auxiliary optimization objective to enhance discrimination during alignment, 
\setlength{\abovedisplayskip}{5pt}
\setlength{\belowdisplayskip}{5pt}
\begin{eqnarray}\label{eq:eq2}\small
	\begin{split}
		\mathcal{L}_{CA}(f_{\mathrm{aug}}^k)\!=\!\sum_{i=1}^{n_{Q_k}}\! \mathcal{H}(\mathcal{S}(f_{\mathrm{aug}}^k(f_{V}(x_i^{(Q_k)})) \cdot f_L(\mathcal{T}_{C}^{k,i})), y_i^{(Q_k)}).
	\end{split}
\end{eqnarray}
Here $\mathcal{H}$ is cross-entropy and $\mathcal{S}$ is softmax.

Eq.(\ref{eq:eq2}) provides standard CLIP supervision during extended domains learning to retain vital class-specific knowledge by clarifying the classification capabilities of the extended domain. Rather than using the composition of the target-domain description and the class name (\textit{i.e.}, $\mathcal{T}_P \! \circ \! \mathcal{T}_C^{k,i}$), we utilize isolated class names (\textit{i.e.}, $\mathcal{T}_C^{k,i}$), aspiring to impart domain invariance and bolster generalization for the extended domains. The extended domains devoid of domain idiosyncrasies facilitates robust transfer to novel distributions, allowing the model to perform well not only in the unseen target domain. We refer to Eq.(\ref{eq:eq1}) as the domain alignment loss and to Eq(\ref{eq:eq2}) as the class alignment loss.
\vspace{-1em}
\paragraph{Distribution Consistency.} 

While the above Domain-Class Alignment induces convergence of the source embeddings towards the unseen target domain, inspection of Eq.(\ref{eq:eq1}) reveals it merely enforces orientation alignment, without constraining the embedding magnitudes. Furthermore, each extended domain has the potential to include class-independent information derived from its respective source domain. Consequently, this fails to mitigate distributional divergence across the resulting augmented domains from disparate sources (Fig.\ref{fig:onecol}(c)), yielding decision boundaries not robust to examples proximate to the data manifold boundaries. We add a distribution consistency loss which force the distribution alignment of the extended domains and remove class-irrelevant information.

The Wasserstein distance provides a geometrically principled metric for comparing probability distributions. Central to its computation is the cost function, which prescribes the expense of transferring unit mass between all locations in the two distributions. 
We denote $\{\bar{Q}_{k}\}_{k=1}^N$ as the true probability distribution of the extended domains. With $\{\Omega_k\}_{k=1}^N$ represents the sample space of $\{\bar{Q}_{k}\}_{k=1}^N$, and $\{\bar{\mathbf{x}}_k\}_{k=1}^N$ represents sample drawn from $\{\Omega_k\}_{k=1}^N$. Therefore, the Wasserstein distance between distributions $\bar{Q}_{i}$ and $\bar{Q}_{j}$ is expressed mathematically as
\setlength{\abovedisplayskip}{5pt}
\setlength{\belowdisplayskip}{5pt}
\begin{eqnarray}\label{eq:eq3}\small
	\begin{split}
		\mathcal{W}(\bar{Q}_{i}, \bar{Q}_{j}) \overset{\mathrm{def}}{=}& \underset{\mathcal{J}\in \Pi (\bar{Q}_{i}, \bar{Q}_{j})}{\mathrm{inf} }  \int_{\Omega_i \times \Omega_j} C(\bar{\mathbf{x}}_i,\bar{\mathbf{x}}_j) d \mathcal{J}(\bar{\mathbf{x}}_i,\bar{\mathbf{x}}_j) \\
		=&  \underset{\mathcal{J}\in \Pi (\bar{Q}_{i}, \bar{Q}_{j})}{\mathrm{inf} }   \left \langle \mathcal{C},\mathcal{J} \right \rangle _{F},
	\end{split}
\end{eqnarray}
where $\mathcal{J} \in \Pi(\bar{Q}_{i}, \bar{Q}_{j})$ denotes a joint probability measure of $\Pi(\bar{Q}_{i}, \bar{Q}_{j})$ with marginals $\bar{Q}_{i}$ and $\bar{Q}_{j}$, $\left \langle \cdot ,\cdot \right \rangle _{F}$ is the Frobenius inner product of matrices, and $\Omega_i \times \Omega_j$ represents the Cartesian product of the sample spaces corresponding to the two probability distributions. $C(\cdot,\cdot)$ is the cost function. $\mathcal{C}$ is the cost matrix, where $\mathcal{C}_{ij} = C(\bar{\mathbf{x}}_i,\bar{\mathbf{x}}_j)$. 
Eq.(\ref{eq:eq3}) states that we only need to sample one joint distribution $\mathcal{J}$ that minimizes the equation, and the minimum value obtained acts as the Wasserstein distance.
In traditional MSDA, commonly used cost functions is Minkowski distance cost, \textit{i.e.}, $C(\bar{\mathbf{x}}_i,\bar{\mathbf{x}}_j) = \left \| \bar{\mathbf{x}}_i-\bar{\mathbf{x}}_j \right \|_p^p$. However, these cost functions do not consider the cost of inter-class transport. Intuitively, the cost of sample transport within and between classes should be different. To address this problem, a few existing works~\cite{flamary2016optimal, montesuma2021wasserstein} incorporate the labels into the cost function. However, these approaches still have limitations as they treat the cost of transport between different classes equally. For instance, transport \textit{dog} to \textit{wolf} incurs the same cost as \textit{dog} to \textit{bird}. However, CLIP's text embeddings situate \textit{dog} and \textit{wolf} in proximity, while positioning \textit{dog} and \textit{bird} more distantly. This uniform treatment of transport cost between classes ultimately impedes the alignment of image and text embeddings. Therefore, we propose a cost function computation method tailored for visual-language models,
\setlength{\abovedisplayskip}{5pt}
\setlength{\belowdisplayskip}{5pt}
\begin{eqnarray}\label{eq:eq4}\small
	\begin{split}
	C(\bar{\mathbf{x}}_i,\bar{\mathbf{x}}_j)\!=\!\mathrm{exp} \big (\frac{\| \bar{\mathbf{x}}_i-\bar{\mathbf{x}}_j  \|^2\!+\!\lambda  \| f_L(\mathcal{T}_C^{i,\cdot})\!-\!f_L(\mathcal{T}_C^{j,\cdot})  \|^2 }{2\sigma^2} \big ).
	\end{split}
\end{eqnarray}

Here $\mathcal{T}_C^{i,\cdot}$ and $\mathcal{T}_C^{j,\cdot}$ denotes the class name corresponding to $\bar{\mathbf{x}}_i$ and $\bar{\mathbf{x}}_j$, respectively. 

The cost function $C(\bar{\mathbf{x}}_i,\bar{\mathbf{x}}_j)$ is defined as a multivariate Gaussian distribution to enable weighted smoothing of cost between sample points from disparate distributions, enhancing robustness against outliers and noise during distance calculations. $\lambda > 0$ controls the degree of emphasis placed on inter-class discrepancies, and $\sigma$ controls the Gaussian width, modulating the smoothness of the distance computations. $C(\bar{\mathbf{x}}_i,\bar{\mathbf{x}}_j)$ incorporates considerations for varying inter-class transfer cost.

In Eq(\ref{eq:eq3}), the solution is guaranteed to be at some vertex in the feasible set $\Pi(\bar{Q}_{i}, \bar{Q}_{j})$, indicating that the solution is a sparse matrix, leading to a highly imbalanced transportation plan. We add an entropy regularization term to address this issue. A larger entropy value indicates higher uncertainty in the solution, which leads to a closer approximation to a uniform distribution and a less sparse solution. Importantly, by introducing the entropy regularization, the original problem can be solved approximately using the Sinkhorn algorithm~\cite{cuturi2013sinkhorn}, significantly reducing the computational cost. Thus, the distribution consistency loss is
\setlength{\abovedisplayskip}{5pt}
\setlength{\belowdisplayskip}{5pt}
\begin{eqnarray}\label{eq:eq5}\small
	\begin{split}
	\mathcal{L}_{DC}(f_{\mathrm{aug}}^k) = \sum_{j=1, j \ne k}^{N} \mathcal{W}(\bar{Q}_{k}, \bar{Q}_{j}) - \frac{1}{\zeta} \mathcal{H}_{\mathrm{reg}}(\mathcal{J}).
	\end{split}
\end{eqnarray}
Here entropy regularization $\mathcal{H}_{\mathrm{reg}}(\mathcal{J}) = -\sum_{i}\mathcal{J}_i \log (\mathcal{J}_i ) $, and $\zeta$ is a hyperparameter.
As the first stage in LanDA, our final objective $f_{\textrm{aug}}^k$ for training the domain-specific augmenter network combines the domain-class alignment loss and distribution consistency loss:
\setlength{\abovedisplayskip}{5pt}
\setlength{\belowdisplayskip}{5pt}
\begin{eqnarray}\label{eq:eq6}\small
	\begin{split}
	\mathcal{L}_{DCA}(f_{\mathrm{aug}}^k)\!=\!\alpha \mathcal{L}_{DA}(f_{\mathrm{aug}}^k)\!+\!\beta \mathcal{L}_{CA}(f_{\mathrm{aug}}^k)\!+\!\gamma \mathcal{L}_{DC}(f_{\mathrm{aug}}^k).
	\end{split}
\end{eqnarray}
$\alpha$, $\beta$ and $\gamma$ is the hyperparameters.
\vspace{-0.7em}
\begin{theorem} \label{Theorem}
	\textsl{Let $\epsilon_{T}(f_{\mathrm{aug}})$ and $\epsilon_{S}(f_{\mathrm{aug}})$ represent the target domain error and source domain error, respectively. $\{\hat{Q}_k\}_{k=1}^{N}$ denotes the associated empirical measure from $\{\bar{Q}_k\}_{k=1}^{N}$. The kernel function $\varphi(\cdot,\cdot)$ solely depends on the utilized VLFMs. $z$ represents the hypothetical empirical measurement of the unseen target domain, while $\mathcal{T}_C^z$ denotes its corresponding class. Let $\vartheta $ represents the combined error of the ideal hypothesis $f_{\mathrm{aug}}^*$ that minimizes the combined error of $\epsilon_{S}(f_{\mathrm{aug}}) + \epsilon_{T}(f_{\mathrm{aug}})$. $w_i$ is the weight of the $i$-th extended domain. For any $\delta > 0$, $\varsigma^{'} < \sqrt{2}$, and any $f_{\mathrm{aug}}$, the following formula holds:}
\setlength{\abovedisplayskip}{5pt}
\setlength{\belowdisplayskip}{0pt}
	\begin{eqnarray}\label{appdex:t1} \small
		\begin{split}
			&\epsilon_{T}(f_{\mathrm{aug}}) \le \epsilon_{S}(f_{\mathrm{aug}}) + \Big\|\mathbb{E}_{z\sim P}[ \varphi(z,\mathcal{T}_C^z)] \Big \|_\varphi + \vartheta + \quad
		\end{split} 
	\end{eqnarray}
\setlength{\abovedisplayskip}{0pt}
\setlength{\belowdisplayskip}{5pt}
	\begin{eqnarray*}\label{appdex:t1} \footnotesize
		\begin{split}
			{\sum_{k=1}^{N}\sum_{j=k+1}^{N}\!w_k w_j} \big[\mathcal{W}(\hat{Q}_k,\!\hat{Q}_j)\!+\!\exp (\!\frac{(1\!+\!\lambda)\!\log(\!\frac{1}{\delta}\!)}{\sigma^2 n_{Q_k}\varsigma^{'}}\!)\!+\!\exp (\!\frac{(1\!+\!\lambda)\!\log(\!\frac{1}{\delta}\!)}{\sigma^2 n_{Q_j}\varsigma^{'}}\!)\big].
		\end{split} 
	\end{eqnarray*}
\end{theorem}

\vspace{-0.1em}
Theorem.\ref{Theorem} indicates that by minimizing both domain-class alignment loss and distribution consistency loss, we can effectively reduce the error of domain-specific augmenters in the unseen target domain. \emph{For the proof and explanation of Theorem.\ref{Theorem}, refer to the suppl. material.}
\subsection{Training Linear Classifier}
\vspace{-0.5em}
After all the domain-specific augmenters is trained, we train a linear probe $f_C$ on all original image embeddings along with the extended domains. Mathematically,
\setlength{\abovedisplayskip}{5pt}
\setlength{\belowdisplayskip}{5pt}
\begin{eqnarray}\label{eq:eq7}\small
	\begin{split}
		\mathcal{L}_C(f_C) = \sum_{k=1}^{N} &\sum_{i=1}^{n_{Q_k}} \varepsilon  \mathcal{H}(\mathcal{S}(f_C(f_V(x_i^{(Q_k)}))),y_i^{(Q_k)})\\
		+&(1\!-\!\varepsilon)\mathcal{H}(\mathcal{S}(f_C(f_{\mathrm{aug}}^k(f_V(x_i^{(Q_k)})))),y_i^{(Q_k)}).
	\end{split}
\end{eqnarray}

$\varepsilon$ is a hyperparameter. Our proposed vision model consists of three components, including (a) a visual feature extractor $f_V$ instantiated from the image decoder of VLFMs, (b) $N$ distinct domain-specific augmenters $\{f_{\mathrm{aug}}^k\}_{k=1}^{N}$, and (c) a linear classifier $f_C$. 
We train only domain-specific augmenters and the linear classifier, which have fewer parameters, and freeze the visual feature extractor, which has large parameters, thereby preserving the feature extraction capability of the pre-trained large-scale VLFMs. Our work can also be seen as a novel approach to fine-tuning large models.

\subsection{Aggregated Target Prediction}
In the testing phase, the goal is to accurately classify a given target image $x_i^{P}$. We combine the extended domains to obtain the final prediction. The key challenge lies in determining the weights for each extended domain. For instance, consider the \textit{sunny} domain, which is closer to the \textit{cloudy} domain than the \textit{night} domain. Thus, it is necessary to assign different weights for each source and target to highlight more relevant sources and suppress irrelevant ones. Fortunately, the difference between each source and target can be assessed using the text embedding of the domain name. Ultimately yielding the final prediction:
\setlength{\abovedisplayskip}{5pt}
\setlength{\belowdisplayskip}{0pt}
\begin{eqnarray}\label{eq:eq7}\small
	\begin{split}
		&\mathrm{Prediction}(x_i^{(P)}) = f_C(\sum_{k=1}^{N}  w_k f_{\mathrm{aug}}^k(f_V(x_i^{(P)}))), \quad \mathrm{where}
	\end{split}
\end{eqnarray}	
\setlength{\abovedisplayskip}{0pt}
\setlength{\belowdisplayskip}{0pt}
\begin{eqnarray*}\footnotesize
	\begin{split}
		w_k\!=\!\frac{\mathcal{W}(\bar{Q}^{\mathcal{T}_{Q_k}}, \bar{P}^{\mathcal{T}_{P}})}{\sum_{j=1}^{N}\!\mathcal{W}(\!\bar{Q}^{\mathcal{T}_{Q_j}}\!,\!\bar{P}^{\mathcal{T}_P}\!)},~~\mathrm{and}~~  C\big(\!f_L\!(\mathcal{T}_{Q_k}\!\circ\!\mathcal{T}_{C}^{k,i}),\!f_L\!(\!\mathcal{T}_{P}\!\circ\!\mathcal{T}_{C}^{k,j})\big)\!=\! \\  
	\end{split}
\end{eqnarray*}
\setlength{\abovedisplayskip}{0pt}
\setlength{\belowdisplayskip}{5pt}
\begin{eqnarray*}\footnotesize
	\begin{split}
		\mathrm{exp} \big (\frac{\| f_L\!(\mathcal{T}_{Q_k}\!\circ\!\mathcal{T}_{C}^{k,i})\!-\!f_L\!(\mathcal{T}_{P}\!\circ\!\mathcal{T}_{C}^{k,j})  \|^2\!+\!\lambda  \| f_L\!(\mathcal{T}_C^{k,i})\!-\!f_L\!(\mathcal{T}_C^{k,j})  \|^2 }{2\sigma^2} \big ).
	\end{split}
\end{eqnarray*}		
Here, $\bar{P}^{\mathcal{T}_P}$ is the probability distribution of the VLFMs text embeddings for the combined set of target domain names and class names. Similarly, $\bar{Q}^{\mathcal{T}_{Q_k}}$ represents the probability distribution of the VLFMs text embeddings for the combined set of source domain names and class names.

\begin{figure*}[t]
	\centering
	\vspace{-0.4em}
	\includegraphics[width=1.0\linewidth]{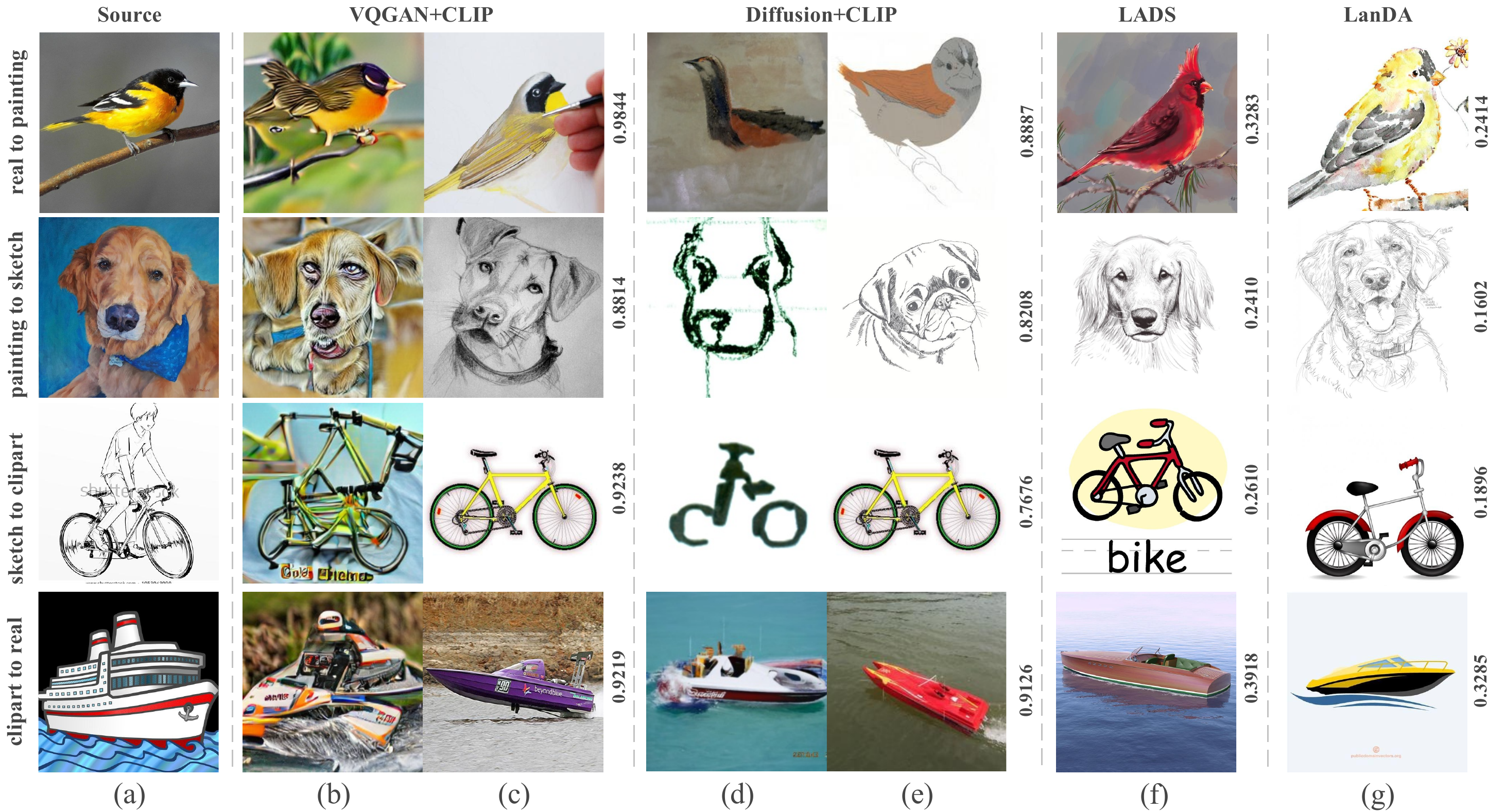}
	\vspace{-1.2em}
	\caption{\textbf{Image generation and nearest neighbor results of different methods in different scenarios.} (a) the source domain images in the order of \textit{real}, \textit{painting}, \textit{sketch} and \textit{clipart}, they are then transferred to \textit{painting}, \textit{sketch}, \textit{clipart} and \textit{real} domains, respectively. (b) and (d) display the target domain image generation results of VQGAN+CLIP and Diffusion+CLIP methods. For both methods, text prompts incorporate the target domain name and class name (e.g., \textit{a painting of a dog}), and image prompts (taken from (a)) are inputted to generate the "target style" images. (c) and (e) present the nearest neighbor results of the generate images within the correct target domain in the CLIP image embedding space. (f) displays the nearest neighbor result obtained from the LADS method within the correct target domain in its augmented embedding space. (g) shows the nearest neighbor result obtained from our method within the correct target domain in the combine extended domain space. We have also indicated the distance of the nearest neighbor.}
	\label{fig:compare}
\end{figure*}
\vspace{-0.4em}
\section{Experiments}
\vspace{-0.15em}
\paragraph{Architecture Detail.} 
For our experiments, we employ three pre-trained vision encoders from CLIP as $f_V$: ViT-L/14, ViT-B/16~\cite{han2022survey} and ResNet-50~\cite{he2016deep} for validation. Here ViT-L/14 boasts a larger parameter count, whereas ViT-B/16 and ResNet-50 have fewer parameters. Additionally, we utilize a transformer-based text encoder as $f_L$. In this section, we present the primary experiments and results conducted on ViT-L/14. We keep all of the visual/text encoder parameters frozen and only update the \textit{domain-specific augmenters} network parameters, each of which is a 2-layer MLP with input and output dimensions of 768 and a hidden dimension of 384. The comprehensive introduction of the datasets, experiments and corresponding discussions regarding ViT-B/16 and ResNet-50, are deferred to \textit{suppl. material}.

\begin{table*}[htbp]
	\footnotesize
	\centering
	\begin{tabularx}{\textwidth}{@{}cXXXXXXXXXXXXXXX@{}}
		\toprule
		\textbf{Scenario}                             & \multicolumn{5}{c}{\textbf{R, P, S to C}}                                                                                                                                              & \multicolumn{5}{c}{\textbf{R, P, S to I}}                                                                                                                                              & \multicolumn{5}{c}{\textbf{R, C, S to P}}                                                                                                                                              \\
		\cmidrule{2-5} \cmidrule{7-10} \cmidrule{12-15}
		\multicolumn{1}{l}{}                          & \multicolumn{3}{c|}{\textbf{ID}}                                       & \multicolumn{1}{c}{\textbf{OD}} & \multicolumn{1}{c}{\cellcolor[HTML]{EFEFEF}}                               & \multicolumn{3}{c|}{\textbf{ID}}                                       & \multicolumn{1}{c}{\textbf{OD}} & \multicolumn{1}{c}{\cellcolor[HTML]{EFEFEF}}                               & \multicolumn{3}{c|}{\textbf{ID}}                                       & \multicolumn{1}{c}{\textbf{OD}} & \multicolumn{1}{c}{\cellcolor[HTML]{EFEFEF}}                               \\
		\multicolumn{1}{l}{\multirow{-2}{*}{\textbf{~~Domains}}} & \multicolumn{1}{c}{R} & \multicolumn{1}{c}{P} & \multicolumn{1}{c}{S} & \multicolumn{1}{c}{C}           & \multicolumn{1}{c}{\multirow{-2}{*}{\cellcolor[HTML]{EFEFEF}\textbf{EXT}}} & \multicolumn{1}{c}{R} & \multicolumn{1}{c}{P} & \multicolumn{1}{c}{S} & \multicolumn{1}{c}{I}           & \multicolumn{1}{c}{\multirow{-2}{*}{\cellcolor[HTML]{EFEFEF}\textbf{EXT}}} & \multicolumn{1}{c}{R} & \multicolumn{1}{c}{C} & \multicolumn{1}{c}{S} & \multicolumn{1}{c}{P}           & \multicolumn{1}{c}{\multirow{-2}{*}{\cellcolor[HTML]{EFEFEF}\textbf{EXT}}} \\
		\midrule
		CLIP ZS(G)~\cite{radford2021learning}                                    & 96.65                 & 94.40                 & 92.54                 & 94.86                           & \cellcolor[HTML]{EFEFEF}94.70                                              & 96.65                 & 94.40                 & 92.54                 & 79.08                           & \cellcolor[HTML]{EFEFEF}86.81                                              & 96.65                 & 94.86                 & 92.54                 & 94.40                           & \cellcolor[HTML]{EFEFEF}94.54                                              
		\\
		CLIP ZS(A)~\cite{radford2021learning}                                    & 96.80                 & 95.32                 & 92.21                 & 94.39                           & \cellcolor[HTML]{EFEFEF}94.58                                              & 96.80                 & \textcolor{blue}{95.32}                 & 92.21                 & 80.90                           & \cellcolor[HTML]{EFEFEF}87.84                                              & 96.80                 & 94.39                 & 92.21                 & 95.32                           & \cellcolor[HTML]{EFEFEF}94.89                                              
		\\
		CLIP LP                                       & 97.21                 & 96.12                 & \textcolor{blue}{95.33}                 & 94.62                           & \cellcolor[HTML]{EFEFEF}95.42                                              &    96.80             &        95.05          &     94.62             &                             80.03    & \cellcolor[HTML]{EFEFEF}    87.76                                               & 97.28                 &  \textcolor{blue}{95.30}                & 95.46                 &   94.16                         & \cellcolor[HTML]{EFEFEF} 95.09
		\\
		VQGAN+CLIP~\cite{crowson2022vqgan}                                    & 97.02                      &\textcolor{blue}{96.30}                       &  94.96                     &    94.65                             & \cellcolor[HTML]{EFEFEF}  95.37                                                 &   96.91                    &     95.30                  & 94.68                      &  80.13                               & \cellcolor[HTML]{EFEFEF}  87.88                                                 &      97.40                 & 94.98                      & 95.64                      &94.93                                 & \cellcolor[HTML]{EFEFEF}   95.47                                                \\
		Diffusion+CLIP~\cite{kim2021diffusionclip}                                     &  97.18                     &  96.27                     &95.24                       & 94.71                                & \cellcolor[HTML]{EFEFEF} 95.47                                                  &    96.86                   & 95.22                      &  94.53                     &      81.16                           & \cellcolor[HTML]{EFEFEF} 88.35                                                  &   97.18                    &  95.11                     &95.73                       &  95.80                               & \cellcolor[HTML]{EFEFEF}   95.90                                                
		\\
		LADS~\cite{dunlap2022using}                                         &   \textcolor{blue}{97.30}                    & 96.19                      &    95.15                   &            \textcolor{blue}{94.74}                     & \cellcolor[HTML]{EFEFEF}   \textcolor{blue}{\textbf{95.48}}                                                &  \textcolor{blue}{97.04}                     &95.12                       &    \textcolor{blue}{94.70}                   &    \textcolor{blue}{81.22}                             & \cellcolor[HTML]{EFEFEF} \textcolor{blue}{\textbf{88.42}}                                                  &    \textcolor{blue}{97.52}                   &   94.89                    &     \textcolor{blue}{95.74}                  &  \textcolor{blue}{96.00}                               & \cellcolor[HTML]{EFEFEF}    \textcolor{blue}{\textbf{96.03}}                                               
		\\
		LanDA(Ours)                                         & \textcolor{red}{97.49}                 & \textcolor{red}{96.67}                 & \textcolor{red}{95.41}                 & \textcolor{red}{95.28}                           & \cellcolor[HTML]{EFEFEF}\textcolor{red}{\textbf{95.90}}                                              &  \textcolor{red}{97.36}                     &        \textcolor{red}{95.60}               &      \textcolor{red}{95.17}                 &    \textcolor{red}{82.06}                             & \cellcolor[HTML]{EFEFEF}  \textcolor{red}{\textbf{89.05}}                                                 &  \textcolor{red}{97.99}                     &        \textcolor{red}{95.40}               &    \textcolor{red}{95.95}                   &    \textcolor{red}{96.63}                             & \cellcolor[HTML]{EFEFEF}\textcolor{red}{\textbf{96.54}} \\
		\bottomrule                                           
	\end{tabularx}
	\vspace{-0.6em}
	\caption{\textbf{Quantitative evaluation in the three-source domain scenario of the Mini-DomainNet dataset.} The “EXT” column is derived as follows: 50\% of the \textit{average} predictions from the ID domain, added to 50\% of the predicted value from the OD domain.}
	\label{E:DomainNet}
	\vspace{-2em}
\end{table*}

\begin{table*}[htbp]
	\footnotesize
	\centering
	\begin{tabularx}{\textwidth}{@{}cXXXXXXXXXXXXXXX@{}}
		\toprule
		\textbf{Scenario}                             & \multicolumn{5}{c}{\textbf{C, P, R to A}}                                                                                                                                              & \multicolumn{5}{c}{\textbf{A, C, P to R}}                                                                                                                                              & \multicolumn{5}{c}{\textbf{A, C, R to P}}                                                                                                                                              \\
		\cmidrule{2-5} \cmidrule{7-10} \cmidrule{12-15}
		\multicolumn{1}{l}{}                          & \multicolumn{3}{c|}{\textbf{ID}}                                       & \multicolumn{1}{c}{\textbf{OD}} & \multicolumn{1}{c}{\cellcolor[HTML]{EFEFEF}}                               & \multicolumn{3}{c|}{\textbf{ID}}                                       & \multicolumn{1}{c}{\textbf{OD}} & \multicolumn{1}{c}{\cellcolor[HTML]{EFEFEF}}                               & \multicolumn{3}{c|}{\textbf{ID}}                                       & \multicolumn{1}{c}{\textbf{OD}} & \multicolumn{1}{c}{\cellcolor[HTML]{EFEFEF}}                               \\
		\multicolumn{1}{l}{\multirow{-2}{*}{\textbf{~~Domains}}} & \multicolumn{1}{c}{C} & \multicolumn{1}{c}{P} & \multicolumn{1}{c}{R} & \multicolumn{1}{c}{A}           & \multicolumn{1}{c}{\multirow{-2}{*}{\cellcolor[HTML]{EFEFEF}\textbf{EXT}}} & \multicolumn{1}{c}{A} & \multicolumn{1}{c}{C} & \multicolumn{1}{c}{P} & \multicolumn{1}{c}{R}           & \multicolumn{1}{c}{\multirow{-2}{*}{\cellcolor[HTML]{EFEFEF}\textbf{EXT}}} & \multicolumn{1}{c}{A} & \multicolumn{1}{c}{C} & \multicolumn{1}{c}{R} & \multicolumn{1}{c}{P}           & \multicolumn{1}{c}{\multirow{-2}{*}{\cellcolor[HTML]{EFEFEF}\textbf{EXT}}} \\
		\midrule
		CLIP ZS(G)~\cite{radford2021learning}                                    & 73.13                 & 90.96                 & 91.94                 & 84.97                           & \cellcolor[HTML]{EFEFEF}85.16                                              & 84.97                 & 73.13                 & 90.96                 & 91.94                           & \cellcolor[HTML]{EFEFEF}87.48                                              & 84.97                 & 73.13                 & 91.94                 & 90.96                           & \cellcolor[HTML]{EFEFEF}87.15                                              \\
		CLIP ZS(A)~\cite{radford2021learning}                                    & 75.34                 & 87.73                 & 92.10                 & 86.34                           & \cellcolor[HTML]{EFEFEF}85.70                                              & 86.34                 & 75.34                 & 87.73                 & 92.10                           & \cellcolor[HTML]{EFEFEF}87.62                                              & 86.34                 & 75.34                 & 92.10                 & 87.73                           & \cellcolor[HTML]{EFEFEF}86.16                                              \\
		CLIP LP                                       &        \textcolor{blue}{83.89}          &    96.08              &             \textcolor{blue}{95.32}     &                  87.02          & \cellcolor[HTML]{EFEFEF}  89.39                                            &       88.97           &         \textcolor{blue}{82.75}         &             \textcolor{blue}{95.18}     &   92.55                              & \cellcolor[HTML]{EFEFEF}   90.76                                            &  90.34                &  83.59                &          \textcolor{blue}{95.78}        &            92.70               & \cellcolor[HTML]{EFEFEF}      91.30                                       \\
		LADS~\cite{dunlap2022using}                                         &  83.13                     &\textcolor{blue}{96.10}                       &     95.24                  &    \textcolor{blue}{87.71}                             & \cellcolor[HTML]{EFEFEF} \textcolor{blue}{\textbf{89.60}}                               &  \textcolor{blue}{90.07}                     &     82.52                  &   95.03                    &                       \textcolor{blue}{93.86}          & \cellcolor[HTML]{EFEFEF}      \textcolor{blue}{\textbf{91.53}}                                             &  \textcolor{blue}{91.86}                     &          \textcolor{blue}{85.44}             &        95.51               & \textcolor{blue}{93.00}                                & \cellcolor[HTML]{EFEFEF}  \textcolor{blue}{\textbf{91.97}}                                                 
		\\
		LanDA(Ours)                                         & \textcolor{red}{85.27}                 & \textcolor{red}{96.31}                 & \textcolor{red}{94.93}                 & \textcolor{red}{88.83}                           & \cellcolor[HTML]{EFEFEF}\textcolor{red}{\textbf{90.50}}                                              & \textcolor{red}{90.48}                      & \textcolor{red}{85.73}                      &     \textcolor{red}{95.63}                  &                           \textcolor{red}{94.09}      & \cellcolor[HTML]{EFEFEF} \textcolor{red}{\textbf{92.35}}                                                  &\textcolor{red}{94.90}                       &\textcolor{red}{85.34}                       &                \textcolor{red}{95.47}       &                         \textcolor{red}{93.22}        & \cellcolor[HTML]{EFEFEF} \textcolor{red}{\textbf{92.56}}\\
		\bottomrule                                           
	\end{tabularx}
	\vspace{-1em}
	\caption{\textbf{Quantitative evaluation in the three-source domain scenario of the Office-Home dataset.}}
	\label{E:office-home}
	\vspace{-1.7em}
\end{table*}

\vspace{-1.3em}
\paragraph{Baselines.} 
Considering the absence of established approaches for language-guided MSDA, we opt to compare against the following methods: (1) \textbf{CLIP ZS(G)}~\cite{radford2021learning} uses the class name alone as the text prompt (\eg, \textit{car}), then compute the similarity between the image embedding and the text embedding for each category, and the image is predicted to be the class with the highest similarity to its text embedding; (2) \textbf{CLIP ZS(A)}~\cite{radford2021learning} uses the composition of class name and domain name as the text prompt (\eg, \textit{a sketch of a car}); (3) \textbf{CLIP LP} fit a linear classifier on top of the CLIP image embeddings; 
(4) \textbf{VQGAN+CLIP}~\cite{crowson2022vqgan} Using a text prompt and an source domain image, we conduct "style transfer" in order to transfer the source image data to the unseen target domain, thereby augmenting the training data. Subsequently, a linear probe is trained on both the augmented and non-augmented CLIP embeddings. 
(5) \textbf{Diffusion+CLIP}~\cite{kim2021diffusionclip} also uses a text prompt and an image prompt to transfer a source domain style image to a target domain style image, and then train a linear classifier.
(6) \textbf{LADS}~\cite{dunlap2022using} is a language-guided domain generalization method applicable to a single source domain. The LADS method is conducted independently for each source domain to the target domain, followed by averaging the prediction results.
Due to the time-consuming nature of image generation, we limit the comparison of VQGAN+CLIP and Diffusion+CLIP methods to the Mini-DomainNet dataset.
\vspace{-0.3em}



\begin{table}[htbp]
	\footnotesize
	\begin{tabularx}{0.478\textwidth}{lXXXX
			>{\columncolor[HTML]{EFEFEF}}l }
		\toprule
		\multicolumn{1}{l}{}                                   & \multicolumn{3}{c|}{\textbf{ID}}                                       & \multicolumn{1}{c}{\textbf{OD}} & \cellcolor[HTML]{EFEFEF}                               \\
		\multicolumn{1}{l}{\multirow{-2}{*}{\textbf{Methods}}} & \multicolumn{1}{c}{R} & \multicolumn{1}{c}{P} & \multicolumn{1}{c}{S} & \multicolumn{1}{c}{C}           & \multirow{-2}{*}{\cellcolor[HTML]{EFEFEF}\textbf{EXT}} \\
		\midrule
		A)CLIP+$\mathcal{L}_{DA}$                                               &         70.34              &  74.77                     &  70.86                     &   75.93                              &           73.96                                             \\
		B)w/ $\mathcal{L}_{CA}$                                                 &  97.35                     &  96.26                     & 95.22                      &  94.80                               &   95.54                                                     \\
		C)w/ $\mathcal{L}_{DC}$                                       &     97.40                  &   96.58                    &   95.39                    &   95.11                              &              95.78                                          \\
		D)w/ LP {\scriptsize (\textit{FULL})}                                               & \textbf{97.49}                 & \textbf{96.67}                 & \textbf{95.41}                 & \textbf{95.28}                           & \textbf{95.90}                    \\
		\bottomrule               
	\end{tabularx}
	\vspace{-0.7em}
	\caption{\textbf{Ablation study in the R,P,S to C scenario of the Mini-DomainNet dataset.}}
	\label{ablation}
	\vspace{-2.0em}
\end{table}

\subsection{Quantitative Evaluation}
\vspace{-0.4em}
Quantitative results on the two benchmark datasets are shown in Table.\ref{E:DomainNet} and Table.\ref{E:office-home}. CLIP LP exhibits satisfactory performance on in-domain (ID), but limited performance on out-of-domain (OD). Table.\ref{E:DomainNet} includes additional pixel-level augmentation baselines, \textit{i.e.}, VQGAN+CLIP and Diffusion+CLIP. However, despite their utilization of CLIP's language knowledge to generate unseen target domain images, the experimental results indicate that Diffusion+CLIP performs better than VQGAN+CLIP on OD. 
Compared with the image generation baselines, LADS has better performance and less time-consuming training. 
The proposed approach LanDA significantly improves over prior state-of-the-art methods in terms of all metrics. Notably, it outperforms other comparative methods not only in the unseen target domain but also exhibits significant improvement in the source domain. This highlights the superiority of our approach in achieving better performance across all domains.

\vspace{-0.4em}
\subsection{Qualitative Evaluation}
\vspace{-0.4em}
Fig.\ref{fig:compare} presents the visualization results obtained from different methods. The generated images of VQGAN+CLIP not only exhibit poor quality but also fail to transfer to the target domain in certain case (e.g., from \textit{painting} to \textit{sketch}), thereby resulting in poor performance on OD.
In contrast, the images generated by the Diffusion+CLIP are closer to the correct target domain and have better performance than the VQGAN+CLIP method on OD. However, it is worth noting that the Diffusion+CLIP method requires significant time and computing resources. Furthermore, we observed that the distance between the images generated by VQGAN+CLIP and Diffusion+CLIP, and the nearest neighbor samples from the correct target domain, still remains considerable in the CLIP image embedding space. 
Both LADS and the proposed method, LanDA, exhibit strong performance in the target domain without the need to generate additional images. However, the nearest neighbor samples of LADS contain more class-independent information, such as the word "\textit{bike}" in the bicycle picture and the water in the speedboat picture  in Fig.\ref{fig:compare}(f). In contrast, LanDA focuses more on domain-invariant features.
\vspace{-0.7em}
\subsection{Ablation Study} 
\vspace{-0.5em}
An ablation study on the R,P,S to C scenario of the Mini-DomainNet Dataset is conducted to validate the effectiveness of various components proposed in the framework of LanDA. In Table.\ref{ablation}, we present comparisons between our full method and its variants:
A) using only the domain alignment loss $\mathcal{L}_{DA}$ to train domain-specific augmenters;
B) using both $\mathcal{L}_{DA}$ and the class alignment loss $\mathcal{L}_{CA}$ to train domain-specific augmenters;
C) using $\mathcal{L}_{DA}$, $\mathcal{L}_{CA}$, and the distribution consistency loss $\mathcal{L}_{DC}$ to train domain-specific augmenters, without training linear classifier, then compute the consistency loss between $\sum_{k=1}^{N} \omega_k f_{\mathrm{aug}}^k(f_V(x_i^{(P)}))$ and the text embedding $f_L(\mathcal{T}_C)$ to obtain prediction results;
D) our full method to train both domain-specific augmenters and a linear classifier.

\begin{figure}[htbp]
	\centering
	\includegraphics[width=1.0\linewidth]{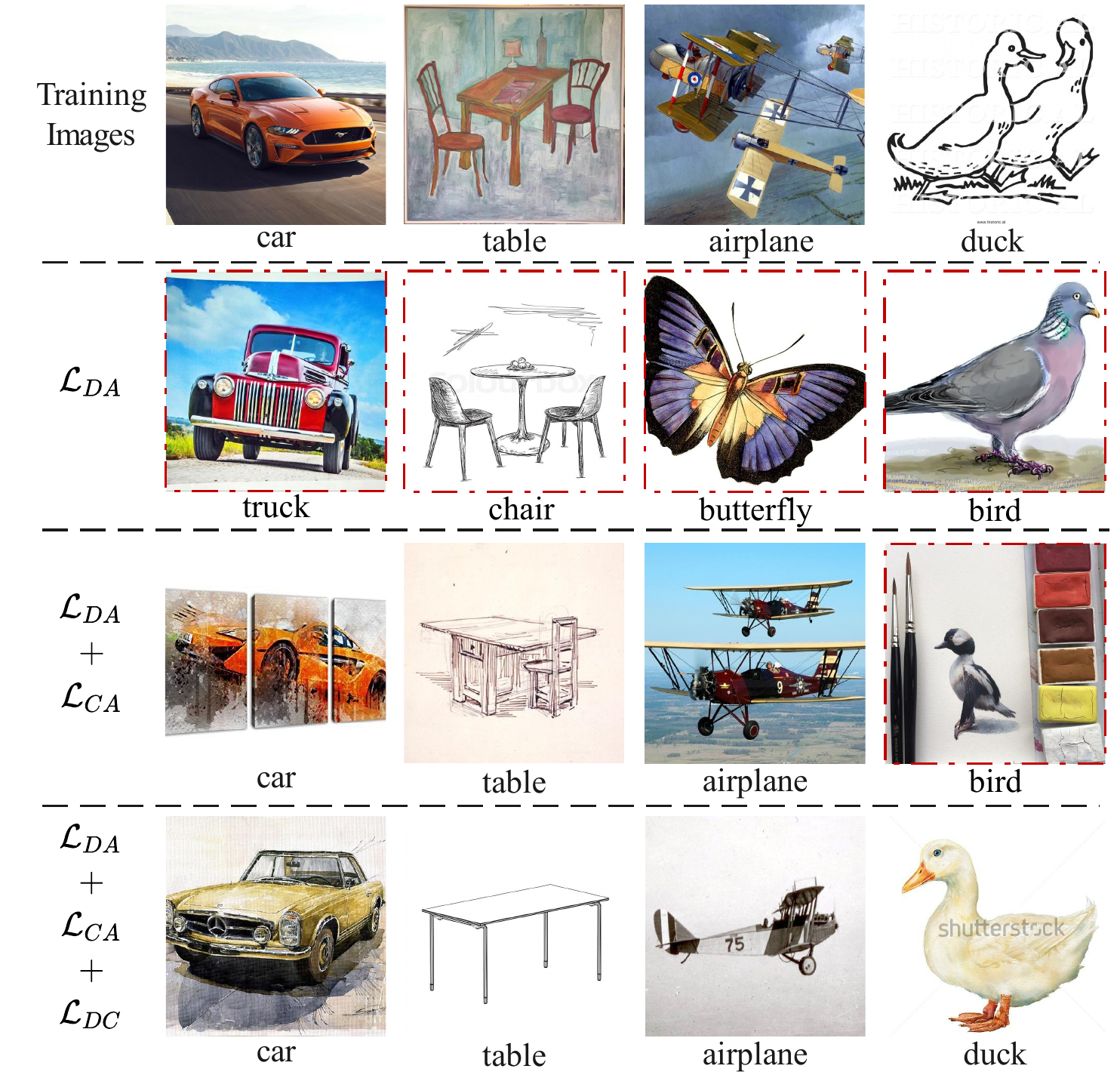}
	\vspace{-1.7em}
	\caption{\textbf{Nearest Neighbors for LanDA when ablating the loss.} The labels below each image represent their respective categories. Images outlined in red have been augmented to error categories. Solely training using $\mathcal{L}_{DA}$ easily result in misclassification, transferring the image to the wrong category. Even though combining $\mathcal{L}_{DA}$ and $\mathcal{L}_{CA}$ helps reduce classification errors, it still contains class-independent information. Our comprehensive method specializes in learning domain-invariant features.}
	\label{fig:ablation}
	\vspace{-1.5em}
\end{figure}

Furthermore, we conduct t-SEN visualizations on the extended domain image embeddings obtained from methods A, B, and C, corresponding to (a), (b), and (c) in Fig.\ref{fig:onecol}, respectively. As depicted in Fig.\ref{fig:onecol}(a), when training the network using only $\mathcal{L}_{DA}$, the class clusters of the extended domains are close to each other, leading to a subpar classification outcome. When employing both $\mathcal{L}_{DA}$ and $\mathcal{L}_{CA}$ during network training, the obtained class clusters become more dispersed, indicating improved classification effectiveness. However, similar to Fig.\ref{fig:onecol}(a), the three extended domain distributions are situated in different positions in Fig.\ref{fig:onecol}(b). This could be attributed to the fact that the extended domains contain class-independent information from their respective source domains. While one domain-specific augmenter performs well for a target distribution located at the edge, the other two perform poorly, resulting in the underutilization of the multi-source domain property. 
In Fig.\ref{fig:onecol}(c), when utilizing our comprehensive loss for training the domain-specific augmenters, it is evident that the boundaries of the extended domains distribution are not distinct, and the class clusters are dispersed across different regions. It means that for a target sample distribution, multiple extended domains are available to validate it, allowing for the effective utilization of the multi-source domain information. The nearest neighbor results presented in Fig.\ref{fig:ablation} demonstrate that $\mathcal{L}_{DC}$ is highly effective in eliminating class-irrelevant information.
\vspace{-1.7em}
\section{Conclusion}
\vspace{-0.5em}
We propose LanDA, a language-guided multi-source domain adaptation method. We view LanDA as a jumping-off point for further exploration on leveraging the powerful capabilities of large-scale multimodal models to improve multi-source domain adaptation accuracy when only language descriptions are provided as input. The method consists of two stages. In the first stage, multiple domain-specific augmenters are trained to align each source domain to the target domain separately in the VLFMs' image and text embedding spaces. Subsequently, the extended domains and class-specific text embeddings are projected into the Wasserstein space for further alignment, extracting domain-invariant information and eliminating class-irrelevant information. 
In the second stage, we merge multiple source domain image embeddings with multiple extended domains to train a shared classifier. The framework comprises a VLFMs vision encoder, multiple domain-specific augmenters, and a linear classifier. During testing, we assign weights by measuring the distance between the text embedding of the source domain and the unseen target domain. Experimental results demonstrate that LanDA achieves strong performance not only in the target domain  but also exhibits improvements in the source domains.

\clearpage
{
	\small
	\bibliographystyle{ieeenat_fullname}
	\bibliography{main}
}
\clearpage
\setcounter{page}{1}
\appendix
\onecolumn
\nolinenumbers
\setlength{\parindent}{0pt}
\rule{\linewidth}{3pt}
\begin{center}
	\Large\bf Supplementary Material for \\
	"LanDA: Language-Guided Multi-Source Domain Adaptation"
\end{center}
\rule{\linewidth}{1.5pt} %
\vspace{-1.5em}
\section{Theoretical Analysis} 
\vspace{-0.5em}
In this section, we introduce error bounds for the proposed LanDA method. Firstly, we decompose the cost function proposed in LanDA and derive a lemma that establishes the relationship between the Wasserstein metric and the error functions of the source and target. Subsequently, we demonstrate how the target error can be bounded for empirical measures.
\vspace{-0.1em}
\begin{definition} \label{definition1}
\textit{Consider $x$ and $z$ as images from distinct source domain, with $\mathcal{T}_C^x$ and $\mathcal{T}_C^z$ representing the textual descriptions of their respective class name. Let $\bar{x}$ and $\bar{z}$ denote their corresponding image embeddings. Since the image and text embeddings output by CLIP are \textbf{one-dimensional}, and dimension $m = $768, 512, or 1024, the following equation holds}
\begin{eqnarray*}\label{appdex:d1}
	\begin{split}
		&C(\bar{x}, \bar{z}) = \mathrm{exp} \big (\frac{\| \bar{x}-\bar{z}  \|^2 + \lambda  \| f_L(\mathcal{T}_C^{x})- f_L(\mathcal{T}_C^z)  \|^2 }{2\sigma^2} \big )\\
		=&\mathrm{exp}(\frac{ {\textstyle \sum_{i}^{m}{\bar{x}}_i^2} }{2\sigma^2} )\mathrm{exp}(\frac{ {\textstyle \sum_{i}^{m}{\bar{z}}_i^2} }{2\sigma^2} )\mathrm{exp}(-\frac{ {\textstyle \sum_{i}^{m}{\bar{x}}_i{\bar{z}}_i} }{\sigma^2} )\mathrm{exp}(\frac{ {\textstyle \lambda  \sum_{i}^{m}[f_L(\mathcal{T}_C^x)]_i^2} }{2\sigma^2} )\mathrm{exp}(\frac{ {\textstyle \lambda\sum_{i}^{m}[f_L(\mathcal{T}_C^z)]_i^2} }{2\sigma^2} )
		\mathrm{exp}(-\frac{ {\textstyle \lambda\sum_{i}^{m}[f_L(\mathcal{T}_C^x)]_i[f_L(\mathcal{T}_C^z)]_i} }{\sigma^2} )\\
		=&\mathrm{exp}(\frac{ {\textstyle \sum_{i}^{m}{\bar{x}}_i^2} }{2\sigma^2} )\mathrm{exp}(\frac{ {\textstyle \sum_{i}^{m}{\bar{z}}_i^2} }{2\sigma^2} )\prod_{i}^{m}\mathrm{exp}(-\frac{ {\textstyle {\bar{x}}_i{\bar{z}}_i} }{\sigma^2} )\mathrm{exp}(\frac{ {\textstyle \lambda\sum_{i}^{m}[f_L(\mathcal{T}_C^x)]_i^2} }{2\sigma^2} )\mathrm{exp}(\frac{ {\textstyle \lambda\sum_{i}^{m}[f_L(\mathcal{T}_C^z)]_i^2} }{2\sigma^2} )
		\prod_{i}^{m}\mathrm{exp}(-\frac{ {\textstyle \lambda[f_L(\mathcal{T}_C^x)]_i[f_L(\mathcal{T}_C^z)]_i} }{\sigma^2})\\
		=&\big[\mathrm{exp}(\frac{ {\textstyle \sum_{i}^{m}{\bar{x}}_i^2} }{2\sigma^2} )\mathrm{exp}(\frac{ {\textstyle \sum_{i}^{m}{\bar{z}}_i^2} }{2\sigma^2} )\prod_{i}^{m}\sum_{n=0}^{\infty } (\frac{ {\textstyle {(-1)^n\bar{x}}_i^n {\bar{z}}_i^n} }{\sigma^{2n}n!} )\big] \\
		&\cdot\big[\mathrm{exp}(\frac{ {\textstyle \lambda\sum_{i}^{m}[f_L(\mathcal{T}_C^x)]_i^2} }{2\sigma^2} )\mathrm{exp}(\frac{ {\textstyle \lambda\sum_{i}^{m}[f_L(\mathcal{T}_C^z)]_i^2} }{2\sigma^2} )
		\prod_{i}^{m}\sum_{n=0}^{\infty } (\frac{ {\textstyle (-1)^n \lambda ^n[f_L(\mathcal{T}_C^x)]_i^n[f_L(\mathcal{T}_C^z)]_i^n} }{\sigma^{2n}n!})\big]\\
		=&\big[\mathrm{exp}(\frac{ {\textstyle \sum_{i}^{m}{\bar{x}}_i^2} }{2\sigma^2} )\mathrm{exp}(\frac{ {\textstyle \sum_{i}^{m}{\bar{z}}_i^2} }{2\sigma^2} )\sum_{n=0}^{\infty }  
		{\textstyle} \prod_{i}^{m}(-1)^n    
		\prod_{i}^{m}(\frac{ {\textstyle {\bar{x}}_i^n } }{\sigma^{n}\sqrt{n!} } )\prod_{i}^{m}(\frac{ {\textstyle {\bar{z}}_i^n } }{\sigma^{n}\sqrt{n!} } )\big] \\
		&\cdot\big[\mathrm{exp}(\frac{ {\textstyle \sum_{i}^{m}[f_L(\mathcal{T}_C^x)]_i^2} }{2\sigma^2} )\mathrm{exp}(\frac{ {\textstyle \sum_{i}^{m}[f_L(\mathcal{T}_C^z)]_i^2} }{2\sigma^2} ) 
		\sum_{n=0}^{\infty }  {\textstyle} \prod_{i}^{m}(-1)^n   
		\prod_{i}^{m} (\frac{ {\textstyle \sqrt{\lambda } [f_L(\mathcal{T}_C^x)]_i^n} }{\sigma^{n}\sqrt{n!}})
		\prod_{i}^{m} (\frac{ {\textstyle \sqrt{\lambda }  [f_L(\mathcal{T}_C^z)]_i^n} }{\sigma^{n}\sqrt{n!}})\big]\\
		=&\Big\{\sum_{n=0}^{\infty }\Big[\Big(\mathrm{exp}(\frac{ {\textstyle \sum_{i}^{m}{\bar{x}}_i^2} }{2\sigma^2} )\mathrm{exp}(\frac{ {\textstyle \sum_{i}^{m}[f_L(\mathcal{T}_C^x)]_i^2} }{2\sigma^2} )\prod_{i}^{m}(\frac{ {\textstyle {\bar{x}}_i^n } }{\sigma^{n}\sqrt{n!} } )\Big)
		\Big(\sum_{n=0}^{\infty }
		\mathrm{exp}(\frac{ {\textstyle \sum_{i}^{m}{\bar{z}}_i^2} }{2\sigma^2} )  \mathrm{exp}(\frac{ {\textstyle \sum_{i}^{m}[f_L(\mathcal{T}_C^z)]_i^2} }{2\sigma^2} ) 
		\prod_{i}^{m}(\frac{ {\textstyle {\bar{z}}_i^n } }{\sigma^{n}\sqrt{n!} } )\Big)\Big] \Big\}\\
		&\cdot\sum_{n=0}^{\infty }
		\prod_{i}^{m} (\frac{ {\textstyle \sqrt{\lambda } [f_L(\mathcal{T}_C^x)]_i^n} }{\sigma^{n}\sqrt{n!}})
		\prod_{i}^{m} (\frac{ {\textstyle \sqrt{\lambda }  [f_L(\mathcal{T}_C^z)]_i^n} }{\sigma^{n}\sqrt{n!}})\\
		\stackrel{\triangle}{=}&\phi (\varphi(x, \mathcal{T}_C^x),x) \phi ( \varphi(z, \mathcal{T}_C^z), z)
		\sum_{n=0}^{\infty } 
		\prod_{i}^{m} (\frac{ {\textstyle \sqrt{\lambda } [f_L(\mathcal{T}_C^x)]_i^n} }{\sigma^{n}\sqrt{n!}})
		\prod_{i}^{m} (\frac{ {\textstyle \sqrt{\lambda }  [f_L(\mathcal{T}_C^z)]_i^n} }{\sigma^{n}\sqrt{n!}})\\
		=&\sum_{n=0}^{\infty }  \Big[\Big(\phi (\varphi(x, \mathcal{T}_C^x),x) \prod_{i}^{m} (\frac{ {\textstyle \sqrt{\lambda } [f_L(\mathcal{T}_C^x)]_i^n} }{\sigma^{n}\sqrt{n!}})\Big) \Big(\sum_{n=0}^{\infty } 
		\phi (\varphi (z, \mathcal{T}_C^z), z)
		\prod_{i}^{m} (\frac{ {\textstyle \sqrt{\lambda }  [f_L(\mathcal{T}_C^z)]_i^n} }{\sigma^{n}\sqrt{n!}}) \Big) \Big]   \\
		%
		%
		%
		\stackrel{\triangle}{=}&\Phi(\phi(\varphi (x, \mathcal{T}_C^x), x), \mathcal{T}_C^x) \Phi(\phi(\varphi (z, \mathcal{T}_C^z), z), \mathcal{T}_C^z) \\
	\end{split}
\end{eqnarray*}
\end{definition}
After normalizing, we can make $\bar{x}_i$, $\bar{z}_i$, $[f_L(\mathcal{T}_C^x)]_i$ and $[f_L(\mathcal{T}_C^z)]_i$ smaller than 1, so that $0 \le \Phi(\phi(\varphi (x, \mathcal{T}_C^x), x), \mathcal{T}_C^x) \le \phi(\varphi (x, \mathcal{T}_C^x), x) \le \varphi (x, \mathcal{T}_C^x) \le 1$. Furthermore, it is evident that $\varphi (x, \mathcal{T}_C^x)$ satisfies the conditions of Mercer's theorem and can therefore be considered a kernel function. We denote the Hebraic space generated by $\varphi (x, \mathcal{T}_C^x)$ as $\left \langle \right \rangle_\varphi$.
\vspace{-0.1em}
\begin{definition} \label{definition2}
\textit{Let's assume that the probability of inconsistency between the image embeddings and the text embeddings of the source domains is}
\begin{eqnarray*}\label{appdex:d2_1}
	\begin{split}
		\epsilon_{S}(f_{\mathrm{aug}},f_{L}) &= \sum_{k=1}^{N} w_k \epsilon_{S_k}(f^k_{{\mathrm{aug}}},f_{L}) =  \sum_{k=1}^{N} w_k \mathbb{E}_{x\sim Q_k}[error(f_{\mathrm{aug}}(f_V(x)),f_{L}(\mathcal{T}_C^x))] 
		= \sum_{k=1}^{N} w_k \mathbb{E}_{x\sim Q_k}[\left \langle \varphi (x, \mathcal{T}_C^x), error\right \rangle _\varphi].
	\end{split}
\end{eqnarray*}
We define the target error in the same manner:
\begin{eqnarray*}\label{appdex:d2_2}
	\begin{split}
		\epsilon_{T}(f_{\mathrm{aug}},f_{T}) = \mathbb{E}_{z\sim P}[error(f_{\mathrm{aug}}(f_V(z)),f_{T}(\mathcal{T}_C^z))] = \mathbb{E}_{z\sim P}[\left \langle \phi (z, \mathcal{T}_C^z), error\right \rangle _\varphi].
	\end{split}
\end{eqnarray*}
\end{definition}

\vspace{-0.1em}
When the source and target error functions are defined w.r.t. $f_{\mathrm{aug}}$ and $f_{L}$, we use the shorthand $\epsilon_{S}(f_{\mathrm{aug}},f_{L}) = \epsilon_{S}(f_{\mathrm{aug}})$.
\begin{lemma} \label{lemma}
\textit{Let cost function $C(\bar{\mathbf{x}}_k,\bar{\mathbf{x}}_j) = \mathrm{exp} \big (\frac{\| \bar{\mathbf{x}}_k-\bar{\mathbf{x}}_j  \|^2 + \lambda  \| f_L(\mathcal{T}_C^{k,\cdot})- f_L(\mathcal{T}_C^{j,\cdot})  \|^2 }{2\sigma^2} \big ) $, and $ \bar{Q}_k, \bar{Q}_j \in \mathcal{P}(\Omega)$ be two probability measures on $\mathbb{R}^m$. For every hypothesis $f_{\mathrm{aug}}^{'}$, $f_{\mathrm{aug}}$, the following holds}
\begin{eqnarray*}\label{appdex:l11}
	\begin{split}
		\epsilon_{T}(f_{\mathrm{aug}},f_{\mathrm{aug}}^{'}) \le \epsilon_{S}(f_{\mathrm{aug}},f_{\mathrm{aug}}^{'}) +  \Big \|
		\mathbb{E}_{z\sim P}[ \varphi(z,\mathcal{T}_C^z)] \Big \|_\varphi 	 + 
		{\sum_{k=1}^{N}\sum_{j=k+1}^{N}w_k w_j} 
		\mathcal{W}(\bar{Q}_k, \bar{Q}_j) .
	\end{split} 
\end{eqnarray*}
\end{lemma}
	
\vspace{-1.5em}
\paragraph{\textit{Proof.}} For simplicity, we assume that $\| error \|_\varphi$ is bounded by 1. This assumption can be verified by imposing appropriate bounds on the norms of $f_{\text{aug}}$ and $f_T$. Furthermore, it can be easily extended to cases where $\| error \|_\varphi \le M$ by scaling, as explained in~\cite{mansour2009domain}.
As the above Lemma.\ref{lemma} plays a pivotal role in the subsequent section, we present its proof here,
\begin{eqnarray*}\label{appdex:p1} 
	\begin{split}
		&~~~~~\epsilon_{T}(f_{\mathrm{aug}},f_{\mathrm{aug}}^{'})
		=  \epsilon_{T}(f_{\mathrm{aug}},f_{\mathrm{aug}}^{'}) -  {\epsilon_{S}(f_{\mathrm{aug}},f_{\mathrm{aug}}^{'})} +  \epsilon_{S}(f_{\mathrm{aug}},f_{\mathrm{aug}}^{'}) \\
		&= \epsilon_{S}(f_{\mathrm{aug}},f_{\mathrm{aug}}^{'}) +
		\mathbb{E}_{z\sim P}[\big \langle \varphi(z,\mathcal{T}_c^z), error\big \rangle _\varphi]  - 
		{\sum_{k=1}^{N} w_k \mathbb{E}_{x\sim Q_k}[\big \langle \varphi(x,\mathcal{T}_C^x), error \big \rangle _\varphi}] \\
		&= \epsilon_{S}(f_{\mathrm{aug}},f_{\mathrm{aug}}^{'}) +  \big \langle
		\mathbb{E}_{z\sim P}[ \varphi(z,\mathcal{T}_C^z)]  - 
		{\sum_{k=1}^{N} w_k \mathbb{E}_{x\sim Q_k}[\varphi(x,\mathcal{T}_C^x)]}, error \big \rangle _\varphi \\
		&\le \epsilon_{S}(f_{\mathrm{aug}},f_{\mathrm{aug}}^{'}) +  \Big \| error  \Big \|_\varphi \Big \|
		\mathbb{E}_{z\sim P}[ \varphi(z,\mathcal{T}_C^z)]  - 
		{\sum_{k=1}^{N} w_k \mathbb{E}_{x\sim Q_k}[\varphi(x,\mathcal{T}_C^x)]} \Big \| _\varphi \\
		&\le \epsilon_{S}(f_{\mathrm{aug}},f_{\mathrm{aug}}^{'}) +  \Big \|
		\mathbb{E}_{z\sim P}[ \varphi(z,\mathcal{T}_C^z)]  - 
		{\sum_{k=1}^{N} w_k \mathbb{E}_{x\sim Q_k}[\varphi(x,\mathcal{T}_C^x)]} \Big \| _\varphi \\
		&\le \epsilon_{S}(f_{\mathrm{aug}},f_{\mathrm{aug}}^{'}) +  \Big \|
		\mathbb{E}_{z\sim P}[ \varphi(z,\mathcal{T}_C^z)]  - 
		{\sum_{k=1}^{N} w_k \mathbb{E}_{x\sim Q_k}[\Phi(\phi(\varphi(x,\mathcal{T}_C^x),x), \mathcal{T}_C^x)]} \Big \| _\varphi \\
		&\le \epsilon_{S}(f_{\mathrm{aug}},f_{\mathrm{aug}}^{'}) +  \Big \|
		\mathbb{E}_{z\sim P}[ \varphi(z,\mathcal{T}_C^z)]  - 
		{\sum_{k=1}^{N} w_k \mathbb{E}_{x\sim Q_k}[\Phi(\phi(\varphi(x,\mathcal{T}_C^x),x), \mathcal{T}_C^x)]}
		{\sum_{j=k+1}^{N} w_j \mathbb{E}_{x\sim Q_j}[\Phi(\phi(\varphi(x,\mathcal{T}_C^x),x), \mathcal{T}_C^x)]}
		\Big \| _\varphi \\
		&= \epsilon_{S}(f_{\mathrm{aug}},f_{\mathrm{aug}}^{'}) +  \Big \|
		\mathbb{E}_{z\sim P}[ \varphi(z,\mathcal{T}_C^z)]  \\
		& \quad\quad\quad\quad\quad\quad\quad  - 
		{\sum_{k=1}^{N}\sum_{j=k+1}^{N}w_k w_j} 
		\int_{\Omega \times \Omega } [\Phi(\phi(\varphi({{\mathbf{x}}_k},\mathcal{T}_C^{\mathbf{x}_k}),{\mathbf{x}}_k), \mathcal{T}_C^{{\mathbf{x}}_k }) ]
		[\Phi(\phi(\varphi({{\mathbf{x}}_j},\mathcal{T}_C^{\mathbf{x}_j}),{\mathbf{x}}_j), \mathcal{T}_C^{{\mathbf{x}}_j}) ]~d\mathcal{J}(\bar{\mathbf{x}}_k,\bar{\mathbf{x}}_j)
		\Big \| _\varphi \\
		&\le \epsilon_{S}(f_{\mathrm{aug}},f_{\mathrm{aug}}^{'}) +  \Big \|
		\mathbb{E}_{z\sim P}[ \varphi(z,\mathcal{T}_C^z)] \Big \|_\varphi \\
		& \quad\quad\quad\quad\quad\quad\quad  + \Big \|
		{\sum_{k=1}^{N}\sum_{j=k+1}^{N}w_k w_j} 
		\int_{\Omega \times \Omega } [\Phi(\phi(\varphi({{\mathbf{x}}_k},\mathcal{T}_C^{{\mathbf{x}}_k}),{\mathbf{x}}_k), \mathcal{T}_C^{{\mathbf{x}}_k }) ]
		[\Phi(\phi(\varphi({{\mathbf{x}}_j},\mathcal{T}_C^{{\mathbf{x}}_j}),{\mathbf{x}}_j), \mathcal{T}_C^{{\mathbf{x}}_j}) ]~d\mathcal{J}(\bar{\mathbf{x}}_k,\bar{\mathbf{x}}_j)
		\Big \| _\Phi \\
		&\le \epsilon_{S}(f_{\mathrm{aug}},f_{\mathrm{aug}}^{'}) +  \Big \|
		\mathbb{E}_{z\sim P}[ \varphi(z,\mathcal{T}_C^z)] \Big \|_\varphi \\
		& \quad\quad\quad\quad\quad\quad\quad  + 
		{\sum_{k=1}^{N}\sum_{j=k+1}^{N}w_k w_j} 
		\int_{\Omega \times \Omega}
		\Big \| [\Phi(\phi(\varphi({{\mathbf{x}}_k},\mathcal{T}_C^{{\mathbf{x}}_k}),{\mathbf{x}}_k), \mathcal{T}_C^{{\mathbf{x}}_k})  ] [\Phi(\phi(\varphi({{\mathbf{x}}_j},\mathcal{T}_C^{{\mathbf{x}}_j}),{\mathbf{x}}_j), \mathcal{T}_C^{{\mathbf{x}}_j})  ]\Big \|_\Phi ~d\mathcal{J}(\bar{\mathbf{x}}_k,\bar{\mathbf{x}}_j)\\
		&\le \epsilon_{S}(f_{\mathrm{aug}},f_{\mathrm{aug}}^{'}) +  \Big \|
		\mathbb{E}_{z\sim P}[ \varphi(z,\mathcal{T}_C^z)] \Big \|_\varphi 	 + 
		{\sum_{k=1}^{N}\sum_{j=k+1}^{N}w_k w_j} 
		\int_{\Omega \times \Omega}C(\bar{\mathbf{x}}_k,\bar{\mathbf{x}}_j)
		~d\mathcal{J}(\bar{\mathbf{x}}_k,\bar{\mathbf{x}}_j)\\
		&= \epsilon_{S}(f_{\mathrm{aug}},f_{\mathrm{aug}}^{'}) +  \Big \|
		\mathbb{E}_{z\sim P}[ \varphi(z,\mathcal{T}_C^z)] \Big \|_\varphi 	 + 
		{\sum_{k=1}^{N}\sum_{j=k+1}^{N}w_k w_j} 
		\mathcal{W}(\bar{Q}_k, \bar{Q}_j) .
	\end{split} 
\end{eqnarray*}

\vspace{-1em}
The second line is obtained by utilizing the reproducing property applied to the function $error(\cdot,\cdot)$. The third line follows from the properties of the expected value. Furthermore, the fourth, ninth, and tenth lines arise from the properties of the inner-product. We note that the fifth line holds due to the earlier discussion that $\| error \| \le 1$. Additional lines are derived through scaling and equivalent substitution. We present the Lemma.\ref{lemma} that introduces Wasserstein distance to relate the source and target error functions in a shared space. 

\vspace{0.15em}
The convergence guarantee of this theorem can be further strengthened.
To establish the validity of our subsequent theorem, we use $\hat{Q}_k = \frac{1}{n_{Q_k}}\sum_{i=1}^{n_{Q_k}}\delta_{i}$ denote the empirical measure to its corresponding true measure $\bar{Q}_k$, with respect to the Wasserstein metric. $\delta_i$ is the sampling probability of the $i$-th sample. The work~\cite{bolley2007quantitative} show that the empirical measure $\hat{Q}_k$ and its true related measure $\bar{Q}_k$ converge in finite samples. For traditional Wasserstein distance $\mathcal{W}_1$, there is an inequality
\begin{eqnarray*}\label{appdex:d0}
 \mathcal{W}_1(\bar{Q}_k, \hat{Q}_k) \le \sqrt{2 \log (\frac{1}{\delta}) / n_{Q_k}\varsigma^{'}},
\end{eqnarray*}
holds for any $\varsigma^{'} < \sqrt{2}$ and $\delta > 0$. In light of this inequality, the upper convergence bound associated with the cost function proposed by us can be deduced straightforwardly.
\begin{eqnarray*}\label{appdex:d1} 
	\begin{split}
		 %
		 \mathcal{W}(\bar{Q}_k, \hat{Q}_k) \le \exp ( \frac{(1+\lambda)\log(1/\delta)}{\sigma^2 n_{Q_k}\varsigma^{'}} ),
	\end{split} 
\end{eqnarray*}
We now utilize it in conjunction with the Lemma.\ref{lemma} to establish the following theorem.

\begin{theorem} \label{Theorem2}
\textsl{Under the assumptions of Lemma.\ref{lemma}, suppose we have $N$ source domains, let $\{\bar{Q}_k\}_{k=1}^{N}$ represent the probability distribution of image embedding, with $\{n_{Q_k}\}_{k=1}^N$ denoting the corresponding number of samples. Let $\{\hat{Q}_k\}_{k=1}^{N}$ denote the associated empirical measure, where $\hat{Q}_k=\frac{1}{n_{Q_k}}\sum_{ i=1}^{n_{Q_k}}\delta_{i}$, and $\delta_i$ is the sampling probability of the $i$-th sample. For any $\varsigma^{'} < \sqrt{2}$, $\delta > 0$, and any $f_{\mathrm{aug}}$, the following formula holds:}
\begin{eqnarray*}\label{appdex:t1} 
	\begin{split}
		\epsilon_{T}(f_{\mathrm{aug}}) \le \epsilon_{S}(f_{\mathrm{aug}})\!+\!\Big\|\mathbb{E}_{z\sim P}[ \varphi(z,\mathcal{T}_c^z)] \Big \|_\varphi\!+\!\Big ({\sum_{k=1}^{N}\sum_{j=k}^{N}w_k w_j} [\mathcal{W}(\hat{Q}_k, \hat{Q}_j)\!+\!\exp ( \frac{(1\!+\!\lambda)\log(1/\delta)}{\sigma^2 n_{Q_k}\varsigma^{'}} )\!+\!\exp ( \frac{(1\!+\!\lambda)\log(1/\delta)}{\sigma^2 n_{Q_j}\varsigma^{'}} ) ] \Big)\!+\!\vartheta,
	\end{split} 
\end{eqnarray*}
\textsl{where $\vartheta $ is the combined error of the ideal hypothesis $f_{\mathrm{aug}}^*$ that minimizes the combined error of $\epsilon_{S}(f_{\mathrm{aug}}) + \epsilon_{T}(f_{\mathrm{aug}})$.} 
\end{theorem}
\vspace{-1em}
\paragraph{\textit{Proof.}} By performing the term shift operation on the formula of Lemma.\ref{lemma}, we obtain $\epsilon_{T}(f_{\mathrm{aug}},f_{\mathrm{aug}}^{'}) - \epsilon_{S}(f_{\mathrm{aug}},f_{\mathrm{aug}}^{'}) \le \Big \|
\mathbb{E}_{z\sim P}[ \varphi(z,\mathcal{T}_c^z)] \Big \|_\varphi 	 + 
{\sum_{k=1}^{N}\sum_{j=k+1}^{N}w_k w_j} 
\mathcal{W}(\bar{Q}_k, \bar{Q}_j)$.
For clarity and ease of expression, we use $\epsilon_{S}(f_{\mathrm{aug}})$ and $\epsilon_{T}(f_{\mathrm{aug}})$ as a representation of $\epsilon_{S}(f_{\mathrm{aug}}, f_L)$ and $\epsilon_{T}(f_{\mathrm{aug}}, f_L)$, avoiding any potential confusion.
\begin{eqnarray*}\label{appdex:d1} 
	\begin{split}
		\epsilon_{T}(f_{\mathrm{aug}}) &\le \epsilon_{T}(f_{\mathrm{aug}}^*) + \epsilon_{T}(f_{\mathrm{aug}}^*, f_{\mathrm{aug}}) \\
		&=  \epsilon_{T}(f_{\mathrm{aug}}^*) + \epsilon_{S}(f_{\mathrm{aug}}, f_{\mathrm{aug}}^*) + \epsilon_{T}(f_{\mathrm{aug}}^*, f_{\mathrm{aug}}) - \epsilon_{S}(f_{\mathrm{aug}}, f_{\mathrm{aug}}^*) \\
		&\le \epsilon_{T}(f_{\mathrm{aug}}^*) + \epsilon_{S}(f_{\mathrm{aug}},f_{\mathrm{aug}}^{*}) + \Big\|\mathbb{E}_{z\sim P}[ \varphi(z,\mathcal{T}_C^z)] \Big \|_\varphi 	 + 
		{\sum_{k=1}^{N}\sum_{j=k+1}^{N}w_k w_j} 
		\mathcal{W}(\bar{Q}_k, \bar{Q}_j)\\
		&\le \epsilon_{T}(f_{\mathrm{aug}}^*) + \epsilon_{S}(f_{\mathrm{aug}}) + \epsilon_{S}(f_{\mathrm{aug}}^{*}) + \Big\|\mathbb{E}_{z\sim P}[ \varphi(z,\mathcal{T}_C^z)] \Big \|_\varphi + {\sum_{k=1}^{N}\sum_{j=k+1}^{N}w_k w_j} 
		\mathcal{W}(\bar{Q}_k, \bar{Q}_j) \\
		&=  \epsilon_{S}(f_{\mathrm{aug}}) + \Big\|\mathbb{E}_{z\sim P}[ \varphi(z,\mathcal{T}_C^z)] \Big \|_\varphi + {\sum_{k=1}^{N}\sum_{j=k+1}^{N}w_k w_j} 
		\mathcal{W}(\bar{Q}_k, \bar{Q}_j) + \vartheta \\
		&\le  \epsilon_{S}(f_{\mathrm{aug}}) + \Big\|\mathbb{E}_{z\sim P}[ \varphi(z,\mathcal{T}_C^z)] \Big \|_\varphi +  {\sum_{k=1}^{N}\sum_{j=k+1}^{N}w_k w_j} (\mathcal{W}(\bar{Q}_k, \hat{Q}_k) + \mathcal{W}(\hat{Q}_k, \bar{Q}_j))  + \vartheta\\
		&\le \epsilon_{S}(f_{\mathrm{aug}}) + \Big\|\mathbb{E}_{z\sim P}[ \varphi(z,\mathcal{T}_C^z)] \Big \|_\varphi + {\sum_{k=1}^{N}\sum_{j=k+1}^{N}w_k w_j} \big(\mathcal{W}(\hat{Q}_k, \hat{Q}_j) + \mathcal{W}(\hat{Q}_j, \bar{Q}_j) + \exp ( \frac{(1\!+\!\lambda)\log(1/\delta)}{\sigma^2 n_{Q_k}\varsigma^{'}} )\big)  + \vartheta \\
		&\le \epsilon_{S}(f_{\mathrm{aug}})\!+\!\Big\|\mathbb{E}_{z\sim P}[ \varphi(z,\mathcal{T}_c^z)] \Big \|_\varphi\!+\!\Big ({\sum_{k=1}^{N}\sum_{j=k+1}^{N}w_k w_j} [\mathcal{W}(\hat{Q}_k, \hat{Q}_j)\!+\!\exp ( \frac{(1\!+\!\lambda)\log(1/\delta)}{\sigma^2 n_{Q_k}\varsigma^{'}} )\!+\!\exp ( \frac{(1\!+\!\lambda)\log(1/\delta)}{\sigma^2 n_{Q_j}\varsigma^{'}} )] \Big)\!+\!\vartheta .\\
	\end{split} 
\end{eqnarray*}

\paragraph{Remark.} Based on the theoretical analysis, the conclusion is evident. To optimize the domain-specific augmenters, it is sufficient to focus solely on optimizing the formula on the right side of the proposed theory. In other words, our objective is to minimize (1) $\epsilon_{S}(f_{\mathrm{aug}})$, which corresponds to the domain-class alignment loss, \textit{i.e.}, $\mathcal{L}_{DA}$ and $\mathcal{L}_{CA}$, and (2) ${\sum_{k=1}^{N}\sum_{j=k+1}^{N}\omega_k\omega_j}\mathcal{W}(\hat{Q}_k, \hat{Q}_j)$, which corresponds to distribution consistency loss, \textit{i.e.}, $\mathcal{L}_{DC}$. In addition to the aforementioned items, the theory further reveals that the learning bound is influenced by the following underlying factors: (1) Size of source domain samples: the more source samples used for training, the higher the accuracy; (2) The performance of the VLFMs is a crucial aspect. In Theorem.\ref{Theorem2}, we encounter the term $\|\mathbb{E}_{z\sim P}[ \varphi(z,\mathcal{T}_c^z)] \|_\varphi$, which represents the correlation between the image embedding and the text embedding of the target domain. Since both image embeddings and text embeddings are generated by VLFMs, this term is only relevant to the performance of VLFMs. 
The better performance of VLFMs, the lower calculated value of this item, which further reduces the value of the cost function and the calculated Wasserstein distance $\mathcal{W}(\hat{Q}_k, \hat{Q}_j)$, leading to a narrower learning bound.
\section{Implementation Detail and Datasets} 
\paragraph{Implementation Detail.} We conducted training by normalizing all text and image embeddings and resizing all images to $224 \times 224$. Typically, the hyperparameters for training the domain-specific augmenters are set as follows: $\alpha = \beta = 1.0$ and $\gamma = 0.05$. The fixed value for cost function: $\sigma = \lambda = 1.0$. $\varepsilon = 0.1$ for training the linear classifier. 
In our experiments, we used the Adam optimizer with an initial learning rate of 1e-2 to train the domain-specific augmenters. For the Mini-DomainNet dataset, we employed a MultiStepLR scheduler with milestones set at [2, 4, 7, 10, 15, 20, 30, 40] epochs and a gamma of 0.5. For the Office-Home dataset, the milestones were set at [5, 10, 15, 20, 30, 40] epochs. 
When training a linear classifier, we set the initial learning rate to 1e-3, and the milestones are set at [1, 3, 5] epochs with a gamma of 0.1.
We train on four NVIDIA GeForce RTX 3090 Ti GPUs with 24GB VRAM each.

\paragraph{Dataset.}
We evaluated LanDA using two existing cross-class datasets, as follows: (1) \textbf{Mini-DomainNet} is a subset of the original DomainNet~\cite{peng2019moment} dataset existing 53,237 images, which contains the 40 most common classes from 5 domains: \textit{sketch}, \textit{infograph}, \textit{clipart}, \textit{painting} and \textit{real}; (2) \textbf{Office-Home}~\cite{venkateswara2017deep} encompasses 15,500 images spanning 66 categories, comprising four domains: \textit{art}, \textit{clipart}, \textit{product} and \textit{real}. We set up various MSDA transfer scenarios for each datasets independently to assess the efficacy of our proposed approach.
\section{Additional Quantitative Evaluation on ViT-L/14}
Table.\ref{APP:ViT16_3} presents supplementary results for the three-source domains transfer scenario conducted on the Mini-DomainNet dataset.
To further complement the quantitative evaluation outlined in the main body, we conducted transfer learning experiments in a two-source domains scenario by employing ViT-L/14 as the visual backbone within the CLIP framework. The transfer experiments for six scenarios on the Office-Home dataset are outlined in Table.\ref{E:TWO_off_1} and Table.\ref{E:TWO_off_2}. Additionally, the transfer experiments for eight scenarios on the Mini-DomainNet dataset are presented in Table.\ref{E:TWO_NET_1} and Table.\ref{E:TWO_NET_2}.
%

\begin{table*}[!htbp]
	\footnotesize
	\centering
	\begin{tabularx}{\textwidth}{@{}cXXXXXXXXXX@{}}
		\toprule
		\textbf{Scenario} & \multicolumn{5}{c}{\textbf{R, P, C to S}}  & \multicolumn{5}{c}{\textbf{C, P, S to R}}      \\
		\cmidrule{2-5} \cmidrule{7-10} \multicolumn{1}{l}{} & \multicolumn{3}{c|}{\textbf{ID}}    & \multicolumn{1}{c}{\textbf{OD}} & \multicolumn{1}{c}{\cellcolor[HTML]{EFEFEF}}  & \multicolumn{3}{c|}{\textbf{ID}}  & \multicolumn{1}{c}{\textbf{OD}} & \multicolumn{1}{c}{\cellcolor[HTML]{EFEFEF}}   \\
		\multicolumn{1}{l}{\multirow{-2}{*}{\textbf{~~Domains}}} & \multicolumn{1}{c}{R} & \multicolumn{1}{c}{P} & \multicolumn{1}{c}{C} & \multicolumn{1}{c}{S}           & \multicolumn{1}{c}{\multirow{-2}{*}{\cellcolor[HTML]{EFEFEF}\textbf{EXT}}} & \multicolumn{1}{c}{C} & \multicolumn{1}{c}{P} & \multicolumn{1}{c}{S} & \multicolumn{1}{c}{R}           & \multicolumn{1}{c}{\multirow{-2}{*}{\cellcolor[HTML]{EFEFEF}\textbf{~EXT}}} \\
		\midrule
		CLIP ZS(G)~\cite{radford2021learning}  &~~~96.65    &~~~94.40   &~~~94.86    &~~~92.54    & \cellcolor[HTML]{EFEFEF}~~~93.21
		&~~~94.86    &~~~94.40   &~~~92.54    &~~~96.65    & \cellcolor[HTML]{EFEFEF}~~~95.29\\
		CLIP ZS(A)~\cite{radford2021learning}  &~~~96.80    &~~~95.32   &~~~94.39    &~~~92.21    & \cellcolor[HTML]{EFEFEF}~~~93.86
		&~~~94.39    &~~~95.32   &~~~92.21    &~~~96.80    & \cellcolor[HTML]{EFEFEF}~~~95.39\\
		CLIP LP                                &~~~97.23    &~~~96.25  &~~~95.67    &~~~93.53    & \cellcolor[HTML]{EFEFEF}~~~94.96
		&~~~95.73    &~~~96.25   &~~~95.66    &~~~96.61    & \cellcolor[HTML]{EFEFEF}~~~96.25\\
		VQGAN+CLIP~\cite{crowson2022vqgan}     &~~~\textcolor{blue}{97.36}    &~~~\textcolor{blue}{96.27}   &~~~95.82    &~~~93.60    & \cellcolor[HTML]{EFEFEF}~~~95.05 
		&~~~95.90    &~~~96.36   &~~~95.71    &~~~96.65    & \cellcolor[HTML]{EFEFEF}~~~96.32\\
		Diffusion+CLIP~\cite{kim2021diffusionclip}  &~~~97.24    &~~~95.84   &~~~95.87    &~~~93.88    & \cellcolor[HTML]{EFEFEF} ~~~95.10
		&~~~\textcolor{blue}{96.32}    &~~~\textcolor{blue}{96.37}   &~~~\textcolor{blue}{95.75}    &~~~96.86    & \cellcolor[HTML]{EFEFEF}~~~\textcolor{blue}{\textbf{96.50}}\\
		LADS~\cite{dunlap2022using}            &~~~97.11    &~~~95.89   &~~~\textcolor{blue}{96.08}    &~~~\textcolor{blue}{94.02}    & \cellcolor[HTML]{EFEFEF} ~~~\textcolor{blue}{\textbf{95.19}}
		&~~~96.30    &~~~96.14   &~~~95.61    &~~~\textcolor{blue}{96.90}    & \cellcolor[HTML]{EFEFEF}~~~96.46\\
		LanDA(Ours)                            &~~~\textcolor{red}{97.47}    &~~~\textcolor{red}{96.05}   &~~~\textcolor{red}{96.04}    &~~~\textcolor{red}{94.33}    & \cellcolor[HTML]{EFEFEF} \textcolor{red}{\textbf{~~~95.43}}
		&\textcolor{red}{~~~96.23}    &\textcolor{red}{~~~96.46}   &~~~\textcolor{red}{95.79}    &~~~\textcolor{red}{97.02}    & \cellcolor[HTML]{EFEFEF}\textcolor{red}{\textbf{~~~96.59}}\\
		\bottomrule                                           
	\end{tabularx}
	\caption{\textbf{Quantitative evaluation in the three-source domains scenario of the Mini-DomainNet dataset.} Each of domain-specific augmenters is a 2-layer MLP with input and output dimensions of 768 and a hidden dimension of 384. The "EXT" column is derived as follows: 50\% of the average predictions from the ID domain, added to 50\% of the predicted value from the OD domain. "ID" column denotes the source domains, "OD" column denotes the unseen target domain.}
	\label{APP:ViT16_3}
\end{table*}

Consistent with the experimental findings in the main body, CLIP LP and VQGAN+CLIP~\cite{crowson2022vqgan} demonstrate satisfactory performance on the source domain test set, however, they do not exhibit significant improvements on the unseen target domain compared to CLIP zero shot in most cases. Conversely, Diddusion+CLIP~\cite{kim2021diffusionclip}, LADS~\cite{dunlap2022using}, and LandA all achieve superior performance to CLIP zero shot in both the target and source domains. Notably, our method, LandA, manifests as the most effective approach.

\begin{table*}[htbp]
	\footnotesize
	\footnotesize
	\centering
	\begin{tabularx}{\textwidth}{@{}cXXXXXXXXXXXX@{}}
		\toprule
		\textbf{Scenario}  & \multicolumn{4}{c}{\textbf{C, R to A}}  
		& \multicolumn{4}{c}{\textbf{P, R to A}} 
		& \multicolumn{4}{c}{\textbf{A, R to P}} \\
		\cmidrule{2-4} \cmidrule{6-8} \cmidrule{10-12} 
		\multicolumn{1}{l}{}  & \multicolumn{2}{c|}{\textbf{ID}}  & \multicolumn{1}{c}{\textbf{OD}} & \multicolumn{1}{c}{\cellcolor[HTML]{EFEFEF}}  & \multicolumn{2}{c|}{\textbf{ID}}  & \multicolumn{1}{c}{\textbf{OD}} & \multicolumn{1}{c}{\cellcolor[HTML]{EFEFEF}}   & \multicolumn{2}{c|}{\textbf{ID}} & \multicolumn{1}{c}{\textbf{OD}} &\multicolumn{1}{c}{\cellcolor[HTML]{EFEFEF}}  \\
		\multicolumn{1}{l}{\multirow{-2}{*}{\textbf{~~Domains}}} & \multicolumn{1}{c}{C} & \multicolumn{1}{c}{R} & \multicolumn{1}{c}{A}           & \multicolumn{1}{c}{\multirow{-2}{*}{\cellcolor[HTML]{EFEFEF}\textbf{EXT}}} & \multicolumn{1}{c}{P} & \multicolumn{1}{c}{R} & \multicolumn{1}{c}{A}           & \multicolumn{1}{c}{\multirow{-2}{*}{\cellcolor[HTML]{EFEFEF}\textbf{EXT}}} & \multicolumn{1}{c}{A} & \multicolumn{1}{c}{R} & \multicolumn{1}{c}{P}           & \multicolumn{1}{c}{\multirow{-2}{*}{\cellcolor[HTML]{EFEFEF}\textbf{EXT}}} \\
		\midrule
		CLIP ZS(G)~\cite{radford2021learning}  & 73.13  &  91.94 & 84.97  &\cellcolor[HTML]{EFEFEF} 83.75
		&  90.96 & 91.94  & 84.97 & \cellcolor[HTML]{EFEFEF} 88.21
		&  84.97 & 91.94  & 90.96  & \cellcolor[HTML]{EFEFEF}89.71 \\
		CLIP ZS(A)~\cite{radford2021learning}   &  75.34 & 92.10  &  86.34 & \cellcolor[HTML]{EFEFEF} 85.03
		&  87.73 &  92.10 & 86.34  & \cellcolor[HTML]{EFEFEF} 88.13
		& 86.34  & 92.10  & 87.73  & \cellcolor[HTML]{EFEFEF} 88.48 \\
		CLIP LP     & \textcolor{blue}{83.13}  & \textcolor{blue}{94.09}  & 86.62  & \cellcolor[HTML]{EFEFEF}  87.62
		&  \textcolor{blue}{95.63} &  92.33 & 86.07  & \cellcolor[HTML]{EFEFEF} 90.03
		&  90.76 & 94.70  & 91.29  & \cellcolor[HTML]{EFEFEF}  92.01    \\
		LADS~\cite{dunlap2022using}        &  82.97  &  93.94&  \textcolor{blue}{87.42} & \cellcolor[HTML]{EFEFEF} \textcolor{blue}{\textbf{87.94}}
		&  94.94 & \textcolor{blue}{93.06}  & \textcolor{blue}{86.45}   & \cellcolor[HTML]{EFEFEF} \textcolor{blue}{\textbf{90.23}}
		&   \textcolor{blue}{91.03} &  \textcolor{blue}{95.00}  &  \textcolor{blue}{92.61}   & \cellcolor[HTML]{EFEFEF}  \textcolor{blue}{\textbf{92.81}}	\\
		LanDA(Ours)      &  \textcolor{red}{83.05} & \textcolor{red}{93.78} &  \textcolor{red}{88.69} & \cellcolor[HTML]{EFEFEF} \textcolor{red}{\textbf{88.55}}
		&  \textcolor{red}{94.88} &  \textcolor{red}{93.02}  &  \textcolor{red}{86.79} & \cellcolor[HTML]{EFEFEF} \textcolor{red}{\textbf{90.37}}
		&  \textcolor{red}{92.62} &   \textcolor{red}{94.93}  &   \textcolor{red}{93.15} & \cellcolor[HTML]{EFEFEF}  \textcolor{red}{\textbf{93.46}}\\
		\bottomrule                                           
	\end{tabularx}
	\vspace{-0.5em}
	\caption{\textbf{Quantitative evaluation in two-source domains scenario of the Office-Home dataset.} }
	\label{E:TWO_off_1}
\end{table*}

\begin{table*}[htbp]
	\footnotesize
	\footnotesize
	\centering
	\begin{tabularx}{\textwidth}{@{}cXXXXXXXXXXXX@{}}
		\toprule
		\textbf{Scenario}  & \multicolumn{4}{c}{\textbf{C, R to P}}  
		& \multicolumn{4}{c}{\textbf{A, P to R}} 
		& \multicolumn{4}{c}{\textbf{C, P to R}} \\
		\cmidrule{2-4} \cmidrule{6-8} \cmidrule{10-12} 
		\multicolumn{1}{l}{}  & \multicolumn{2}{c|}{\textbf{ID}}  & \multicolumn{1}{c}{\textbf{OD}} & \multicolumn{1}{c}{\cellcolor[HTML]{EFEFEF}}  & \multicolumn{2}{c|}{\textbf{ID}}  & \multicolumn{1}{c}{\textbf{OD}} & \multicolumn{1}{c}{\cellcolor[HTML]{EFEFEF}}   & \multicolumn{2}{c|}{\textbf{ID}} & \multicolumn{1}{c}{\textbf{OD}} &\multicolumn{1}{c}{\cellcolor[HTML]{EFEFEF}}  \\
		\multicolumn{1}{l}{\multirow{-2}{*}{\textbf{~~Domains}}} & \multicolumn{1}{c}{C} & \multicolumn{1}{c}{R} & \multicolumn{1}{c}{P}           & \multicolumn{1}{c}{\multirow{-2}{*}{\cellcolor[HTML]{EFEFEF}\textbf{EXT}}} & \multicolumn{1}{c}{A} & \multicolumn{1}{c}{P} & \multicolumn{1}{c}{R}           & \multicolumn{1}{c}{\multirow{-2}{*}{\cellcolor[HTML]{EFEFEF}\textbf{EXT}}} & \multicolumn{1}{c}{C} & \multicolumn{1}{c}{P} & \multicolumn{1}{c}{R}           & \multicolumn{1}{c}{\multirow{-2}{*}{\cellcolor[HTML]{EFEFEF}\textbf{EXT}}} \\
		\midrule
		CLIP ZS(G)~\cite{radford2021learning}  &  73.13 &  91.94 & 90.96 &\cellcolor[HTML]{EFEFEF} 86.75
		& 84.97  & 90.96  & 91.94 & \cellcolor[HTML]{EFEFEF} 89.95
		& 73.13  & 90.96  & 91.94  & \cellcolor[HTML]{EFEFEF} 86.99 \\
		CLIP ZS(A)~\cite{radford2021learning}   &  75.34 &  92.10 & 87.73  & \cellcolor[HTML]{EFEFEF} 85.73
		& 86.34  & 87.73  & 92.10  & \cellcolor[HTML]{EFEFEF} 89.57
		& 75.34  & 87.73  & 92.10  & \cellcolor[HTML]{EFEFEF} 86.82 \\
		CLIP LP     & \textcolor{blue}{81.53}  &  \textcolor{blue}{94.86} &   90.81& \cellcolor[HTML]{EFEFEF}  89.50
		&  \textcolor{blue}{89.38} & 95.78  & 92.32  & \cellcolor[HTML]{EFEFEF} 92.45
		&  \textcolor{blue}{82.67} & \textcolor{blue}{95.48}  &  92.47 & \cellcolor[HTML]{EFEFEF}   90.62   \\
		LADS~\cite{dunlap2022using}        &  81.44  & 94.51 & \textcolor{blue}{91.27}  & \cellcolor[HTML]{EFEFEF} \textcolor{blue}{\textbf{89.72}}
		& 89.33  & \textcolor{blue}{95.94}  & \textcolor{blue}{93.06}   & \cellcolor[HTML]{EFEFEF} \textcolor{blue}{\textbf{92.85}}
		& 82.10  &  94.96 &  \textcolor{blue}{92.82}  & \cellcolor[HTML]{EFEFEF} \textcolor{blue}{\textbf{90.68}}		\\
		LanDA(Ours)      &  \textcolor{red}{82.06} & \textcolor{red}{94.24} &  \textcolor{red}{91.80} & \cellcolor[HTML]{EFEFEF} \textcolor{red}{\textbf{89.98}}
		& \textcolor{red}{89.69}  &  \textcolor{red}{95.86}  &  \textcolor{red}{93.79} & \cellcolor[HTML]{EFEFEF} \textcolor{red}{\textbf{93.28}}
		&  \textcolor{red}{82.37}   & \textcolor{red}{95.78} &  \textcolor{red}{93.55} & \cellcolor[HTML]{EFEFEF}\textcolor{red}{\textbf{91.31}}\\
		\bottomrule                                           
	\end{tabularx}
	\vspace{-0.5em}
	\caption{\textbf{Quantitative evaluation in two-source domains scenario of the Office-Home dataset.}}
	\label{E:TWO_off_2}
\end{table*}

\begin{table*}[htbp]
	\footnotesize
	\footnotesize
	\centering
	\begin{tabularx}{\textwidth}{@{}cXXXXXXXXXXXXXXXX@{}}
		\toprule
		\textbf{Scenario}  & \multicolumn{4}{c}{\textbf{R, S to C}}  
		& \multicolumn{4}{c}{\textbf{R, P to C}} 
		& \multicolumn{4}{c}{\textbf{R, S to I}}
		& \multicolumn{4}{c}{\textbf{R, P to I}} \\
		\cmidrule{2-4} \cmidrule{6-8} \cmidrule{10-12} \cmidrule{14-16}
		\multicolumn{1}{l}{}  & \multicolumn{2}{c|}{\textbf{ID}}  & \multicolumn{1}{c}{\textbf{OD}} & \multicolumn{1}{c}{\cellcolor[HTML]{EFEFEF}}  & \multicolumn{2}{c|}{\textbf{ID}}  & \multicolumn{1}{c}{\textbf{OD}} & \multicolumn{1}{c}{\cellcolor[HTML]{EFEFEF}}   & \multicolumn{2}{c|}{\textbf{ID}} & \multicolumn{1}{c}{\textbf{OD}} &\multicolumn{1}{c}{\cellcolor[HTML]{EFEFEF}} &
		\multicolumn{2}{c|}{\textbf{ID}}  & \multicolumn{1}{c}{\textbf{OD}} & \multicolumn{1}{c}{\cellcolor[HTML]{EFEFEF}}  \\
		\multicolumn{1}{l}{\multirow{-2}{*}{\textbf{~~Domains}}} & \multicolumn{1}{c}{R} & \multicolumn{1}{c}{S} & \multicolumn{1}{c}{C}           & \multicolumn{1}{c}{\multirow{-2}{*}{\cellcolor[HTML]{EFEFEF}\textbf{EXT}}} & \multicolumn{1}{c}{R} & \multicolumn{1}{c}{P} & \multicolumn{1}{c}{C}           & \multicolumn{1}{c}{\multirow{-2}{*}{\cellcolor[HTML]{EFEFEF}\textbf{EXT}}} & \multicolumn{1}{c}{R} & \multicolumn{1}{c}{S} & \multicolumn{1}{c}{I}           & \multicolumn{1}{c}{\multirow{-2}{*}{\cellcolor[HTML]{EFEFEF}\textbf{EXT}}} & \multicolumn{1}{c}{R} & \multicolumn{1}{c}{P} & \multicolumn{1}{c}{I}           & \multicolumn{1}{c}{\multirow{-2}{*}{\cellcolor[HTML]{EFEFEF}\textbf{EXT}}} \\
		\midrule
		CLIP ZS(G)~\cite{radford2021learning}  &  96.65 &  92.54 &  \textcolor{blue}{94.86} &\cellcolor[HTML]{EFEFEF} 94.73
		& 96.65  & 94.40  & \textcolor{blue}{94.86} & \cellcolor[HTML]{EFEFEF} 95.19
		&  96.65 & 92.54  & 79.08  & \cellcolor[HTML]{EFEFEF} 86.34
		&  96.65 & 94.40  & 79.08  & \cellcolor[HTML]{EFEFEF}  87.30\\
		CLIP ZS(A)~\cite{radford2021learning}   &  96.80 & 92.21  & 94.39  & \cellcolor[HTML]{EFEFEF} 94.45
		&  96.80 & 95.32  & 94.39  & \cellcolor[HTML]{EFEFEF} 95.23
		&  96.80 & 92.21  & 80.90  & \cellcolor[HTML]{EFEFEF} 87.70
		&  96.80 & 95.32  & 80.90  & \cellcolor[HTML]{EFEFEF} 88.48 \\
		CLIP LP     &  97.34 & \textcolor{blue}{96.08}  &  94.49 & \cellcolor[HTML]{EFEFEF}  95.60
		&  97.55 &  96.73 &  94.68 & \cellcolor[HTML]{EFEFEF} 95.91
		&  97.31 &  95.46 &  79.25 & \cellcolor[HTML]{EFEFEF}  87.82
		&  \textcolor{blue}{97.58} &  95.01 & 79.79  & \cellcolor[HTML]{EFEFEF}   88.04    \\
		VQGAN+CLIP~\cite{crowson2022vqgan}         &   97.41 & 95.92 &  94.50 & \cellcolor[HTML]{EFEFEF} 95.58
		&  \textcolor{blue}{97.62} &  96.76 &  94.64  & \cellcolor[HTML]{EFEFEF} 95.92
		&  97.42 & 95.56  &  79.31  & \cellcolor[HTML]{EFEFEF}   87.90  
		&  97.47 & 95.27  &  79.80   & \cellcolor[HTML]{EFEFEF} 	88.09	\\
		Diffusion+CLIP~\cite{kim2021diffusionclip}        &  \textcolor{blue}{97.60}  & 95.83 & 94.68  & \cellcolor[HTML]{EFEFEF} \textcolor{blue}{\textbf{95.70}}
		&  97.56 &  \textcolor{blue}{96.80} &  94.70  & \cellcolor[HTML]{EFEFEF} \textcolor{blue}{\textbf{95.94}}
		&  \textcolor{blue}{97.48} &  95.58 &  79.92  & \cellcolor[HTML]{EFEFEF}  88.23   
		& 97.50  & 95.35  & 80.87 & \cellcolor[HTML]{EFEFEF} 88.65		\\
		LADS~\cite{dunlap2022using}        &  97.48  & 95.90 & 94.64  & \cellcolor[HTML]{EFEFEF} 95.67
		& 97.44   &  96.75 &  94.72  & \cellcolor[HTML]{EFEFEF} 95.91
		& 97.37  & \textcolor{blue}{95.60}  &  \textcolor{blue}{80.94}  & \cellcolor[HTML]{EFEFEF}   \textcolor{blue}{\textbf{88.71}}  
		& 97.34  &  \textcolor{blue}{95.46} &  \textcolor{blue}{81.03}  & \cellcolor[HTML]{EFEFEF} 	\textcolor{blue}{\textbf{88.72}}	\\
		LanDA(Ours)      & \textcolor{red}{97.57}  & \textcolor{red}{95.89} & \textcolor{red}{95.24}  & \cellcolor[HTML]{EFEFEF} \textcolor{red}{\textbf{95.99}}
		& \textcolor{red}{97.49}  &  \textcolor{red}{96.87}  & \textcolor{red}{95.25}  & \cellcolor[HTML]{EFEFEF} \textcolor{red}{\textbf{96.22}}
		& \textcolor{red}{97.49} &  \textcolor{red}{95.71}  &  \textcolor{red}{81.29} & \cellcolor[HTML]{EFEFEF} \textcolor{red}{\textbf{88.95}}
		& \textcolor{red}{97.36}  &  \textcolor{red}{95.70}  &  \textcolor{red}{81.55} & \cellcolor[HTML]{EFEFEF} \textcolor{red}{\textbf{89.04}}      \\
		\bottomrule                                           
	\end{tabularx}
	\caption{\textbf{Quantitative evaluation in two-source domains scenario of the Mini-DomainNet dataset.}}
	\label{E:TWO_NET_1}
\end{table*}

\begin{table*}[htbp]
	\footnotesize
	\footnotesize
	\centering
	\begin{tabularx}{\textwidth}{@{}cXXXXXXXXXXXXXXXX@{}}
		\toprule
		\textbf{Scenario}  & \multicolumn{4}{c}{\textbf{R, C to S}}  
		& \multicolumn{4}{c}{\textbf{R, P to S}} 
		& \multicolumn{4}{c}{\textbf{R, S to P}}
		& \multicolumn{4}{c}{\textbf{R, C to P}} \\
		\cmidrule{2-4} \cmidrule{6-8} \cmidrule{10-12} \cmidrule{14-16}
		\multicolumn{1}{l}{}  & \multicolumn{2}{c|}{\textbf{ID}}  & \multicolumn{1}{c}{\textbf{OD}} & \multicolumn{1}{c}{\cellcolor[HTML]{EFEFEF}}  & \multicolumn{2}{c|}{\textbf{ID}}  & \multicolumn{1}{c}{\textbf{OD}} & \multicolumn{1}{c}{\cellcolor[HTML]{EFEFEF}}   & \multicolumn{2}{c|}{\textbf{ID}} & \multicolumn{1}{c}{\textbf{OD}} &\multicolumn{1}{c}{\cellcolor[HTML]{EFEFEF}} &
		\multicolumn{2}{c|}{\textbf{ID}}  & \multicolumn{1}{c}{\textbf{OD}} & \multicolumn{1}{c}{\cellcolor[HTML]{EFEFEF}}  \\
		\multicolumn{1}{l}{\multirow{-2}{*}{\textbf{~~Domains}}} & \multicolumn{1}{c}{R} & \multicolumn{1}{c}{C} & \multicolumn{1}{c}{S}           & \multicolumn{1}{c}{\multirow{-2}{*}{\cellcolor[HTML]{EFEFEF}\textbf{EXT}}} & \multicolumn{1}{c}{R} & \multicolumn{1}{c}{P} & \multicolumn{1}{c}{S}           & \multicolumn{1}{c}{\multirow{-2}{*}{\cellcolor[HTML]{EFEFEF}\textbf{EXT}}} & \multicolumn{1}{c}{R} & \multicolumn{1}{c}{S} & \multicolumn{1}{c}{P}           & \multicolumn{1}{c}{\multirow{-2}{*}{\cellcolor[HTML]{EFEFEF}\textbf{EXT}}} & \multicolumn{1}{c}{R} & \multicolumn{1}{c}{C} & \multicolumn{1}{c}{P}           & \multicolumn{1}{c}{\multirow{-2}{*}{\cellcolor[HTML]{EFEFEF}\textbf{EXT}}} \\
		\midrule
		CLIP ZS(G)~\cite{radford2021learning}  &  96.65 &  94.86 & 92.54  &\cellcolor[HTML]{EFEFEF} 94.15
		& 96.65  & 94.40  & 92.54 & \cellcolor[HTML]{EFEFEF}   94.03
		&  96.65 & 92.54  & 94.40  & \cellcolor[HTML]{EFEFEF}  94.50
		&  96.65 & 94.86  & 94.40  & \cellcolor[HTML]{EFEFEF}  95.08\\
		CLIP ZS(A)~\cite{radford2021learning}   & 96.80  &  94.39 & 92.21  & \cellcolor[HTML]{EFEFEF} 93.90
		&  96.80 & 95.32  & 92.21  & \cellcolor[HTML]{EFEFEF} 94.14
		&  96.80 & 92.21  & 95.32  & \cellcolor[HTML]{EFEFEF} 94.91
		&  96.80 & 94.39  & 95.32  & \cellcolor[HTML]{EFEFEF} 95.46 \\
		CLIP LP     &  \textcolor{blue}{97.34} & 95.61  & 92.61  & \cellcolor[HTML]{EFEFEF} 94.54 
		& 96.79  & 95.57  &  92.58 & \cellcolor[HTML]{EFEFEF} 94.38
		& \textcolor{blue}{97.10}  & 95.41  & 93.95  & \cellcolor[HTML]{EFEFEF}  95.10
		&  97.45 &  95.98 &  93.88 & \cellcolor[HTML]{EFEFEF}   93.30    \\
		VQGAN+CLIP~\cite{crowson2022vqgan}        &  97.26  & 95.47 & 92.63  & \cellcolor[HTML]{EFEFEF} 94.50
		&  96.81 &  95.48 &  92.60  & \cellcolor[HTML]{EFEFEF} 94.37
		&  96.96 &  \textcolor{blue}{95.50} &   94.87 & \cellcolor[HTML]{EFEFEF}  95.55   
		&  97.26 & 95.90  &  94.84  & \cellcolor[HTML]{EFEFEF} 95.71		\\
		Diffusion+CLIP~\cite{kim2021diffusionclip}        &  97.29  & 95.55 &  93.14 & \cellcolor[HTML]{EFEFEF} 94.78
		&  \textcolor{blue}{96.93} &95.55   & \textcolor{blue}{92.97}   & \cellcolor[HTML]{EFEFEF} \textcolor{blue}{\textbf{94.61}}
		&  97.02 &  95.48 &   95.64 & \cellcolor[HTML]{EFEFEF}  \textcolor{blue}{\textbf{95.95}}   
		& \textcolor{blue}{97.38}  & 96.07  &  95.68  & \cellcolor[HTML]{EFEFEF} 	96.20	\\
		LADS~\cite{dunlap2022using}        &  97.32  & \textcolor{blue}{95.76} &  \textcolor{blue}{93.22} & \cellcolor[HTML]{EFEFEF} \textcolor{blue}{\textbf{94.88}}
		& 96.77  &  \textcolor{blue}{95.60} &  92.84  & \cellcolor[HTML]{EFEFEF} 94.51
		& 96.89  &  95.26 &  \textcolor{blue}{95.68}  & \cellcolor[HTML]{EFEFEF} 95.88    
		&  97.36 &  \textcolor{blue}{96.12} &   \textcolor{blue}{95.79} & \cellcolor[HTML]{EFEFEF} 	\textcolor{blue}{\textbf{96.27}}	\\
		LanDA(Ours)      & \textcolor{red}{97.37} & \textcolor{red}{95.78} &  \textcolor{red}{93.78} & \cellcolor[HTML]{EFEFEF} \textcolor{red}{\textbf{95.18}}
		& \textcolor{red}{96.93}  &  \textcolor{red}{95.46}  &  \textcolor{red}{93.33} & \cellcolor[HTML]{EFEFEF} \textcolor{red}{\textbf{94.76}}
		& \textcolor{red}{97.29}  &  \textcolor{red}{95.79}  &  \textcolor{red}{95.94} & \cellcolor[HTML]{EFEFEF} \textcolor{red}{\textbf{96.24}}
		&  \textcolor{red}{97.29} &  \textcolor{red}{95.98}  & \textcolor{red}{96.10}  & \cellcolor[HTML]{EFEFEF} \textcolor{red}{\textbf{96.38}}      \\
		\bottomrule                                           
	\end{tabularx}
	\caption{\textbf{Quantitative evaluation in two-source domains scenario of the Mini-DomainNet dataset.}}
	\label{E:TWO_NET_2}
\end{table*}

\vspace{-1em}
\section{Additional Quantitative Evaluation on ViT-B/16 and RN50} 
In previous experiments, ViT-L/14 was utilized as the visual backbone of CLIP, employing its larger model size and capability to extract multiple hierarchical features. However, in this section, we assess the effectiveness of ViT-B/16 and RN50 (ResNet50) as alternative backbones. ViT-B/16, compared to ViT-L/14, offers a smaller model size, making it well-suited for real-time applications and providing faster inference speed. Detailed experimental results can be found in Table.\ref{APP:VITB16}.
\begin{table*}[ht]
	\footnotesize
	\centering
	\begin{tabularx}{\textwidth}{@{}cXXXXXXXXXXXXXXX@{}}
		\toprule
		\textbf{Scenario} & \multicolumn{5}{c}{\textbf{Mini-DomainNet: R, P, S to C~~~~~~~~~~~~~~}}  & \multicolumn{5}{c}{\textbf{Mini-DomainNet: R, P, C to S~~~~~~~~~~~~~~}}  & \multicolumn{5}{c}{\textbf{Office-Home: A, C, R to P~~~~~~~~~~~~~~~~~}}     \\
		\cmidrule{2-5} \cmidrule{7-10} \cmidrule{12-15}\multicolumn{1}{l}{} & \multicolumn{3}{c|}{\textbf{ID}}    & \multicolumn{1}{c}{\textbf{OD}} & \multicolumn{1}{c}{\cellcolor[HTML]{EFEFEF}}  & \multicolumn{3}{c|}{\textbf{ID}}  & \multicolumn{1}{c}{\textbf{OD}} & \multicolumn{1}{c}{\cellcolor[HTML]{EFEFEF}}   & \multicolumn{3}{c|}{\textbf{ID}}  & \multicolumn{1}{c}{\textbf{OD}} & \multicolumn{1}{c}{\cellcolor[HTML]{EFEFEF}} \\
		\multicolumn{1}{l}{\multirow{-2}{*}{\textbf{~~Domains}}} & \multicolumn{1}{c}{R} & \multicolumn{1}{c}{P} & \multicolumn{1}{c}{S} & \multicolumn{1}{c}{C}           & \multicolumn{1}{c}{\multirow{-2}{*}{\cellcolor[HTML]{EFEFEF}\textbf{EXT}}} & \multicolumn{1}{c}{R} & \multicolumn{1}{c}{P} & \multicolumn{1}{c}{C} & \multicolumn{1}{c}{S}           & \multicolumn{1}{c}{\multirow{-2}{*}{\cellcolor[HTML]{EFEFEF}\textbf{EXT}}} & \multicolumn{1}{c}{A} & \multicolumn{1}{c}{C} & \multicolumn{1}{c}{R} & \multicolumn{1}{c}{P}           & \multicolumn{1}{c}{\multirow{-2}{*}{\cellcolor[HTML]{EFEFEF}\textbf{EXT}}} \\
		\midrule
		CLIP ZS(G)~\cite{radford2021learning}  &95.85    &93.09   &90.12    &92.39    & \cellcolor[HTML]{EFEFEF}92.70 
		                                       &95.85    &93.09   &92.39    &90.12    & \cellcolor[HTML]{EFEFEF}91.95
		                                       &79.45    &63.13   &87.72    &86.14    & \cellcolor[HTML]{EFEFEF}81.45\\
		CLIP ZS(A)~\cite{radford2021learning}  &96.11    &93.19   &90.83    &92.64    & \cellcolor[HTML]{EFEFEF}93.01 
		                                       &96.11    &93.19   &92.64    &90.83    & \cellcolor[HTML]{EFEFEF}92.01
	                     	                   &78.07    &63.36   &87.34    &88.68    & \cellcolor[HTML]{EFEFEF}82.47\\
		CLIP LP                                &96.53    &94.47   &\textcolor{blue}{93.41}    &92.20    & \cellcolor[HTML]{EFEFEF}93.50 
		                                       &96.52    &94.39   &94.34    &90.89    & \cellcolor[HTML]{EFEFEF}92.99
		                                       &85.24    &75.14   &90.71    &88.75    & \cellcolor[HTML]{EFEFEF}86.22\\
		VQGAN+CLIP~\cite{crowson2022vqgan}     &96.57    &\textcolor{blue}{94.52}   &93.33    &92.28    & \cellcolor[HTML]{EFEFEF}93.54 
		                                       &\textcolor{blue}{96.80}    &\textcolor{blue}{94.68}   &\textcolor{blue}{94.42}    &90.99    & \cellcolor[HTML]{EFEFEF} 93.15
		                                       &\textcolor{gray}{\textit{N.A.}}    &\textcolor{gray}{\textit{N.A.}}   &\textcolor{gray}{\textit{N.A.}}    &\textcolor{gray}{\textit{N.A.}}    & \cellcolor[HTML]{EFEFEF}\textcolor{gray}{\textit{N.A.}}\\
		Diffusion+CLIP~\cite{kim2021diffusionclip}  &\textcolor{blue}{96.66}    &94.51   &93.28    &\textcolor{blue}{92.75}    & \cellcolor[HTML]{EFEFEF} 93.78
		                                            &96.77    &94.25   &94.16    &91.43    & \cellcolor[HTML]{EFEFEF}93.25
		                                            &\textcolor{gray}{\textit{N.A.}}    &\textcolor{gray}{\textit{N.A.}}   &\textcolor{gray}{\textit{N.A.}}    &\textcolor{gray}{\textit{N.A.}}    & \cellcolor[HTML]{EFEFEF}\textcolor{gray}{\textit{N.A.}}\\
		LADS~\cite{dunlap2022using}            &96.55    &94.48   &93.40    &92.69    & \cellcolor[HTML]{EFEFEF}\textcolor{blue}{\textbf{93.75}} 
		                                       &96.58    &94.60   &94.03    &\textcolor{blue}{91.47}    & \cellcolor[HTML]{EFEFEF}\textcolor{blue}{\textbf{93.27}}
		                                       &\textcolor{blue}{85.58}    &\textcolor{blue}{75.79}   &\textcolor{blue}{91.04}    &\textcolor{blue}{89.12}    & \cellcolor[HTML]{EFEFEF}\textcolor{blue}{\textbf{86.63}}\\
		LanDA(Ours)                            &\textcolor{red}{96.60}    &\textcolor{red}{94.33}   &\textcolor{red}{93.45}    &\textcolor{red}{93.03}    & \cellcolor[HTML]{EFEFEF} \textcolor{red}{\textbf{93.91}}
		                                       &\textcolor{red}{96.70}    &\textcolor{red}{94.71}   &\textcolor{red}{94.31}    &\textcolor{red}{91.79}    & \cellcolor[HTML]{EFEFEF}\textcolor{red}{\textbf{93.52}}
		                                       &\textcolor{red}{85.79}    &\textcolor{red}{77.94}   &\textcolor{red}{91.20}    &\textcolor{red}{89.96}    & \cellcolor[HTML]{EFEFEF}\textcolor{red}{\textbf{87.47}}\\
		\bottomrule                                           
	\end{tabularx}
	\caption{\textbf{Quantitative evaluation in the three-source domains scenario.} Each of domain-specific augmenters is a 2-layer MLP with input and output dimensions of 512 and a hidden dimension of 256.}
	\label{APP:VITB16}
\end{table*}

RN50 (ResNet50) is commonly used as the visual backbone in traditional Multi-Source Domain Adaptation (MSDA) methods due to its fewer parameters compared to ViT-L/14 and ViT-B/16. The experimental results, presented in Table.\ref{APP:RN50}, showcase the performance achieved when utilizing ResNet50 as the visual backbone of CLIP.

\begin{table*}[!htbp]
	\footnotesize
	\centering
	\begin{tabularx}{\textwidth}{@{}cXXXXXXXXXXXXXXX@{}}
		\toprule
		\textbf{Scenario} & \multicolumn{5}{c}{\textbf{Mini-DomainNet: R, P, S to C~~~~~~~~~~~~~~}}  & \multicolumn{5}{c}{\textbf{Mini-DomainNet: R, P, C to S~~~~~~~~~~~~~~}}  & \multicolumn{5}{c}{\textbf{Office-Home: A, C, R to P~~~~~~~~~~~~~~~~~}}     \\
		\cmidrule{2-5} \cmidrule{7-10} \cmidrule{12-15}\multicolumn{1}{l}{} & \multicolumn{3}{c|}{\textbf{ID}}    & \multicolumn{1}{c}{\textbf{OD}} & \multicolumn{1}{c}{\cellcolor[HTML]{EFEFEF}}  & \multicolumn{3}{c|}{\textbf{ID}}  & \multicolumn{1}{c}{\textbf{OD}} & \multicolumn{1}{c}{\cellcolor[HTML]{EFEFEF}}   & \multicolumn{3}{c|}{\textbf{ID}}  & \multicolumn{1}{c}{\textbf{OD}} & \multicolumn{1}{c}{\cellcolor[HTML]{EFEFEF}} \\
		\multicolumn{1}{l}{\multirow{-2}{*}{\textbf{~~Domains}}} & \multicolumn{1}{c}{R} & \multicolumn{1}{c}{P} & \multicolumn{1}{c}{S} & \multicolumn{1}{c}{C}           & \multicolumn{1}{c}{\multirow{-2}{*}{\cellcolor[HTML]{EFEFEF}\textbf{EXT}}} & \multicolumn{1}{c}{R} & \multicolumn{1}{c}{P} & \multicolumn{1}{c}{C} & \multicolumn{1}{c}{S}           & \multicolumn{1}{c}{\multirow{-2}{*}{\cellcolor[HTML]{EFEFEF}\textbf{EXT}}} & \multicolumn{1}{c}{A} & \multicolumn{1}{c}{C} & \multicolumn{1}{c}{R} & \multicolumn{1}{c}{P}           & \multicolumn{1}{c}{\multirow{-2}{*}{\cellcolor[HTML]{EFEFEF}\textbf{EXT}}} \\
		\midrule
		CLIP ZS(G)~\cite{radford2021learning}  &93.20    &85.05   &79.41    &80.94    & \cellcolor[HTML]{EFEFEF} 83.41
		&93.20    &85.05   &80.94    &79.41    & \cellcolor[HTML]{EFEFEF}82.90
		&67.59    &45.88   &78.97    &77.71    & \cellcolor[HTML]{EFEFEF}70.93\\
		CLIP ZS(A)~\cite{radford2021learning}  &93.55    &85.94   &79.95    &82.76    & \cellcolor[HTML]{EFEFEF} 84.62
		&93.55    &85.94   &82.67    &79.95    & \cellcolor[HTML]{EFEFEF}83.67
		&61.79    &39.39   &76.98    &77.03    & \cellcolor[HTML]{EFEFEF}68.21\\
		CLIP LP                                &94.94    &\textcolor{blue}{89.10}   &82.67    &82.67    & \cellcolor[HTML]{EFEFEF} 85.79
		&95.30    &\textcolor{blue}{89.21}   &90.22    &80.49    & \cellcolor[HTML]{EFEFEF}86.03
		&\textcolor{blue}{75.86}    &65.11   &85.80    &79.64    & \cellcolor[HTML]{EFEFEF}77.62\\
		VQGAN+CLIP~\cite{crowson2022vqgan}     &\textcolor{blue}{95.01}    &88.95   &82.63    &82.80    & \cellcolor[HTML]{EFEFEF} 85.83
		&\textcolor{blue}{95.36}    &88.89   &90.25    &80.53    & \cellcolor[HTML]{EFEFEF}86.02
		&\textcolor{gray}{\textit{N.A.}}    &\textcolor{gray}{\textit{N.A.}}   &\textcolor{gray}{\textit{N.A.}}    &\textcolor{gray}{\textit{N.A.}}    & \cellcolor[HTML]{EFEFEF}\textcolor{gray}{\textit{N.A.}}\\
		Diffusion+CLIP~\cite{kim2021diffusionclip}  &94.88    &88.97   &83.52    &83.28    & \cellcolor[HTML]{EFEFEF}86.20 
		&95.28    &88.94   &\textcolor{blue}{90.30}    &81.02    & \cellcolor[HTML]{EFEFEF}\textcolor{blue}{\textbf{86.26}}
		&\textcolor{gray}{\textit{N.A.}}    &\textcolor{gray}{\textit{N.A.}}   &\textcolor{gray}{\textit{N.A.}}    &\textcolor{gray}{\textit{N.A.}}    & \cellcolor[HTML]{EFEFEF}\textcolor{gray}{\textit{N.A.}}\\
		LADS~\cite{dunlap2022using}            &94.90    &89.24   &\textcolor{blue}{85.75}    &\textcolor{blue}{83.44}    & \cellcolor[HTML]{EFEFEF} \textcolor{blue}{\textbf{86.70}}
		&95.20    &88.76   &89.88    &\textcolor{blue}{81.14}    & \cellcolor[HTML]{EFEFEF}86.21
		&75.22    &\textcolor{blue}{67.97}    &\textcolor{blue}{86.05}    &\textcolor{blue}{80.20}    & \cellcolor[HTML]{EFEFEF}\textcolor{blue}{\textbf{78.31}}\\
		LanDA(Ours)                            &\textcolor{red}{95.00}    &\textcolor{red}{89.48}   &\textcolor{red}{87.33}    &\textcolor{red}{84.03}    & \cellcolor[HTML]{EFEFEF} \textcolor{red}{\textbf{87.32}}
		&\textcolor{red}{95.38}    &\textcolor{red}{88.83}   &\textcolor{red}{89.79}    &\textcolor{red}{81.66}    & \cellcolor[HTML]{EFEFEF}\textcolor{red}{\textbf{86.50}}
		&\textcolor{red}{75.07}    &\textcolor{red}{69.16}   &\textcolor{red}{86.33}    &\textcolor{red}{80.87}    & \cellcolor[HTML]{EFEFEF}\textcolor{red}{\textbf{78.86}}\\
		\bottomrule                                           
	\end{tabularx}
	\caption{\textbf{Quantitative evaluation in the three-source domains scenario.} Each of domain-specific augmenters is a 2-layer MLP with input and output dimensions of 1024 and a hidden dimension of 512.}
	\label{APP:RN50}
\end{table*}

Upon comparing all experimental results in the main body and the outcomes presented in Table.\ref{E:TWO_off_1} through \ref{E:TWO_NET_2} in the appendix, it is evident that when using ViT-B/16 and RN50 as the visual backbone of CLIP, CLIP LP and VQGAN-CLIP~\cite{crowson2022vqgan} struggle to surpass CLIP ZS~\cite{radford2021learning} in OD in certain cases (\eg, \textit{A, C, R to P} in Table.\ref{APP:VITB16}, and \textit{R, P, S to C} in Table.\ref{APP:RN50}). 
The hypothesized reason is that a reduced parameter count compromises the ability to extract accurate features for each class, particularly when dealing with vast datasets. However, other methods excluding CLIP ZS~\cite{radford2021learning} exhibit substantial improvements in OD performance. The results presented in Table.\ref{APP:VITB16} and Table.\ref{APP:RN50} illustrate that the proposed LanDA method continues to achieve superior outcomes, highlighting the effectiveness of our approach within a lightweight general ViT and CNN architecture.

\section{Additional t-SNE Visualizations}
\begin{figure*}[htbp]
	\centering
	\includegraphics[width=0.95\linewidth]{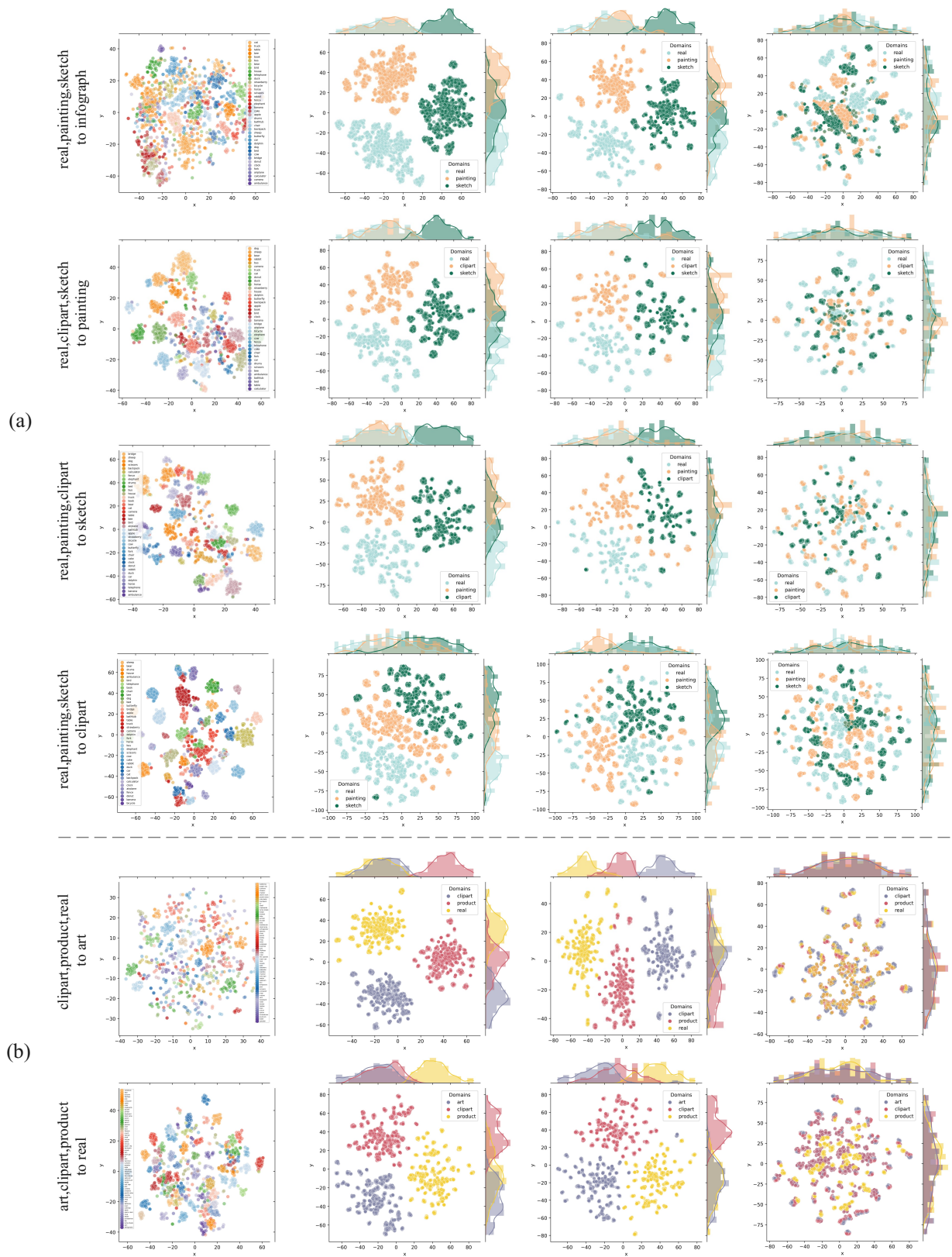}
	\caption{\textbf{t-SNE visualizations on the Mini-DomainNet and Office-Home datasets.}}
	\label{fig:t_sen}
\end{figure*}

Figure.\ref{fig:t_sen} displays the t-SNE visualization results of the ablation experiment, corresponding to Figure 2 in the main body. In Figure.\ref{fig:t_sen}(a), the experiment was conducted on the Mini-DomainNet dataset, while in Figure.\ref{fig:t_sen}(b), it was performed on the Office-Home dataset.
The leftmost column of the figure showcases the visualized image embeddings obtained from CLIP, where each color represents a specific class. The remaining columns illustrate the extended domains, with distinct colors indicating different image embeddings from these domains. Our study compares three approaches for training domain-specific augmenters: the second column utilizes $\mathcal{L}_{DA}$ alone, the third column combines $\mathcal{L}_{DA}$ and $\mathcal{L}_{CA}$, and the fourth column integrates $\mathcal{L}_{DA}$, $\mathcal{L}_{CA}$, and $\mathcal{L}_{DC}$.
Figure.\ref{fig:t_sen} provides an intuitive observation where the second column exhibits a noticeable distance between domains, along with relatively closely clustered points within each domain. In the third column, the clusters within domains are more dispersed, while the inter-domain distance remains substantial. The fourth column demonstrates a close distribution of the three domains, suggesting the extraction of common features, while also showcasing dispersed clusters within each domain, indicating a strong classification ability.

\clearpage
\section{Examples of the Generation Images Produced by VQGAN+CLIP}
\begin{figure*}[thbp]
	\centering
	\includegraphics[width=\linewidth]{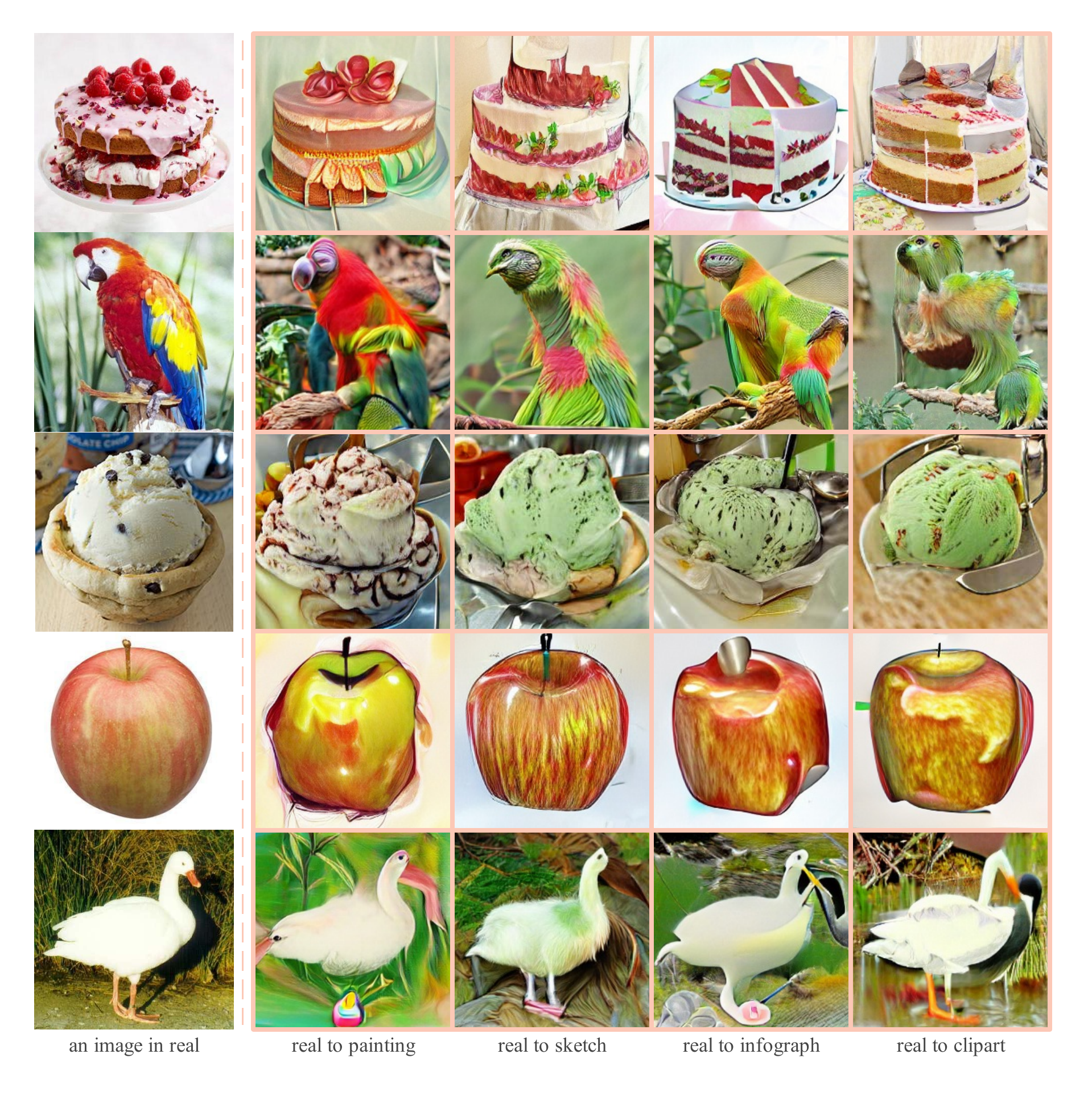}
	
	\caption{\textbf{We give examples from the VQGAN+CLIP method.} Beginning with \textit{a realistic photo of a cake}, \textit{a realistic photo of a parrot}, \textit{a realistic photo of an ice cream}, \textit{a realistic photo of an apple} and \textit{a realistic photo of a swan}. Then we transfer these images into the domains of \textit{painting}, \textit{sketch}, \textit{infograph}, and \textit{clipart}. Simultaneously inputting a text prompt in a desired style and an image prompt from a source domain can be employed to achieve a "style transfer" effect. For each pairing of source and target domains, we systematically generate 200 images, amounting to an aggregate of 3000 images for training. Subsequently, we train a classifier for CLIP using both the source domains data and the generated data.}
	\label{fig:gan}
\end{figure*}

\clearpage
\section{Examples of the Generation Images Produced by Diffusion+CLIP}
\begin{figure*}[htbp]
	\centering
	\includegraphics[width=\linewidth]{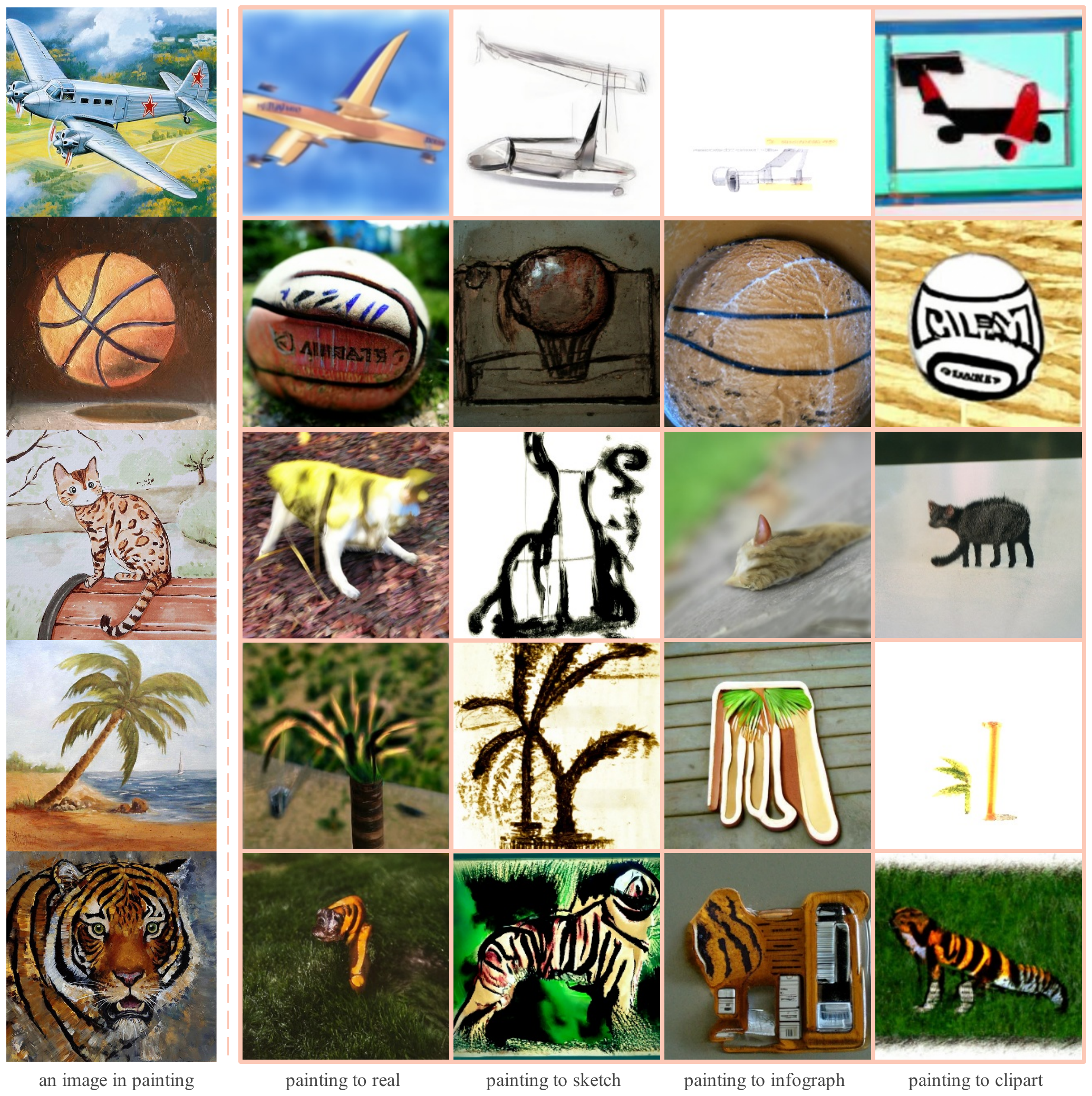}
	\caption{\textbf{We give examples from the Diffusion+CLIP method.} Beginning with \textit{a painting of an airplane}, \textit{a painting of a basketball}, \textit{a painting of a cat}, \textit{a painting of a palm tree} and \textit{a painting of a tiger}. Then we transfer these images into the domains of \textit{real}, \textit{sketch}, \textit{infograph}, and \textit{clipart}. Simultaneously inputting a text prompt describing the target domain and an image prompt of the source domain can promote the transfer of images from the source domain to the target domain. Given the time-consuming nature of the diffusion-based image generation process,  we systematically generate 200 images for each source domain-target domain pairing, thereby accumulating a total of 400 or 600 images for each transfer scenario. Consequently, our experiments yield a comprehensive set of 3000 images. Analogously, we simultaneously feed the source domains and generated target domain images into CLIP to training the classifier.}
	\label{fig:ddpms}
\end{figure*}

\clearpage
\section{Compared with Conventional MSDA Methods} 
To ensure fairness, quantitative comparisons with the traditional MSDA method~\cite{xu2018deep, venkat2020your, peng2019moment} were not conducted in our previous experiments. In this section, we outline the key differences between these two approaches.
\vspace{0.5em}
\begin{itemize}
	\item Traditional MSDA methods, whether source-unfree or source-free, operate under the assumption that the target domain is accessible (see Figure.\ref{fig:compare}(a)). While our method does not rely on any target domain images.
	\vspace{0.3em}
	\item The majority of conventional MSDA methods necessitate training a substantial number of parameters, typically ranging in the tens of millions. But our method only requires training hundreds of thousands to millions of parameters.
	\vspace{0.3em}
	\item Similar to previous experimental baselines, our method incorporates VLFMs as prior knowledge. Additionally, the performance of LanDA is closely tied to the capabilities of VLFMs.
	\vspace{0.3em}
	\item Even when utilizing ResNet50 as the vision backbone, our method can achieve comparable performance to traditional MSDA methods (see Figure.\ref{fig:compare}(b)), even without a single target domain image and using a reduced number of training parameters. The performance of our proposed method significantly outperforms the traditional MSDA method when utilizing ViT-L/14 or ViT-B/16 as the backbone.
	\vspace{0.3em}
	\item Our method demonstrates strong performance in both the target and source domains, whereas traditional methods primarily focus on the performance solely on the target domain.
\end{itemize}
\vspace{0.5em}

\begin{figure*}[htbp]
	\centering
	\includegraphics[width=\linewidth]{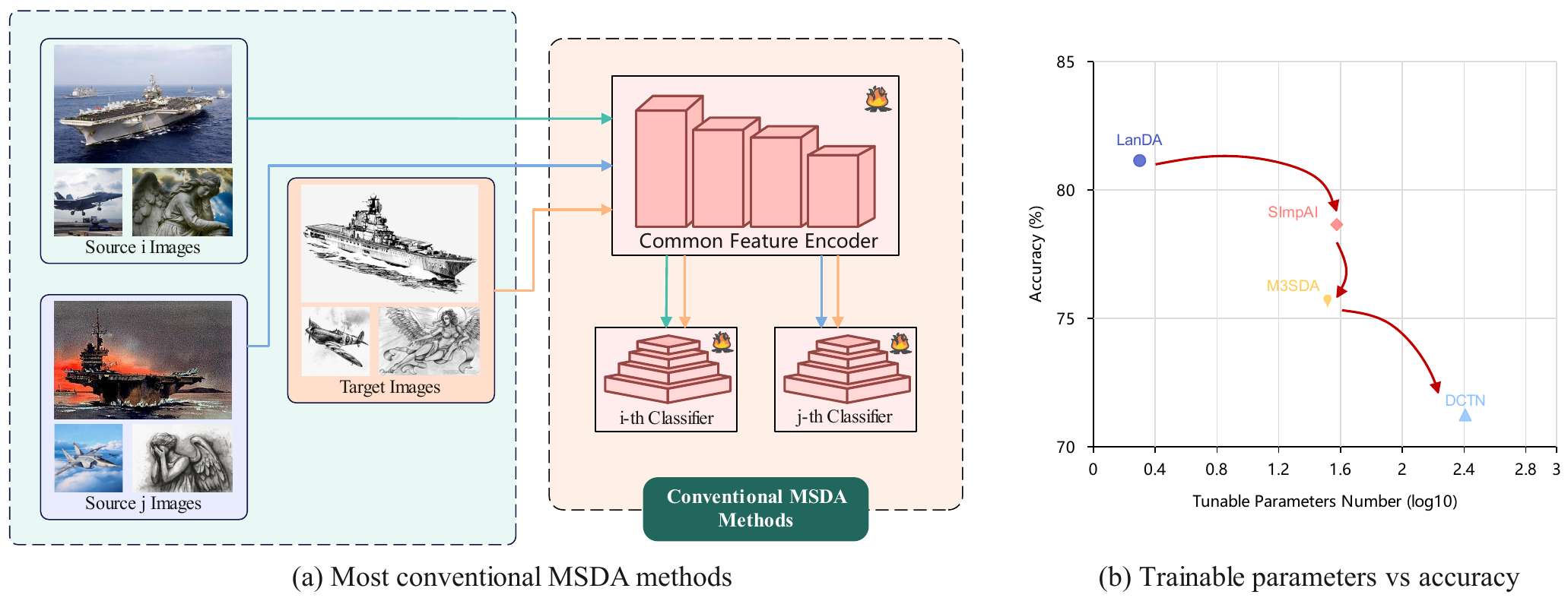}
	\caption{\textbf{Compared with Conventiona MSDA methods.} (a) Most conventional MSDA methods employ a shared feature extractor alongside domain-specific classifier heads, and the target domain images is required. (b) We compare the \textit{A, C, R, $\to$ P} transfer scenario of various methods on the Office-Home dataset when utilizing RN50 as the visual backbone. The trainable parameters of LanDA are significantly less, ranging from 1/8 to 1/5, in comparison to the conventiona MSDA method.}
	\label{fig:compare}
\end{figure*}

\section{Limitations and Future Work} 
Given the propose approach reliance on the rich concepts learned by pre-trained visual-language models, its applicability is constrained to domains that can be accurately depicted through textual descriptions and are sufficiently represented in the data employed for training the pre-trained visual-language models.
Our proposed method has limitations when it comes to domain transformations in complex situations. We hope that future research will facilitate more sophisticated domain transformations that are not reliant on the visual language pre-trained models' accuracy in classifying existing domains and classes.



\end{document}